\pdfoutput=1

\documentclass[11pt]{article}

\usepackage[]{ACL2023}

\usepackage{times}
\usepackage{latexsym}
\usepackage{arydshln} 

\usepackage[T1]{fontenc}

\usepackage[utf8]{inputenc}

\usepackage{microtype}

\usepackage{inconsolata}
\usepackage{booktabs}
\usepackage{graphicx}
\usepackage{soul}
\usepackage[dvipsnames]{xcolor}
\usepackage[most]{tcolorbox}
\usepackage{tabularx}
\usepackage{enumitem}
\usepackage{float}
\usepackage{bbding}
\usepackage{pifont}

\title{Minimal Clips, Maximum Salience: Long Video Summarization \\via Key Moment Extraction}

\author{
\textbf{
Galann Pennec\textsuperscript{$\infty,\diamondsuit,\heartsuit$} 
\ \quad
Zhengyuan Liu\textsuperscript{$\diamondsuit,\heartsuit$} 
\ \quad 
}
\\
\textbf{
Nicholas Asher\textsuperscript{$\S,\heartsuit$} 
\ \quad 
Philippe Muller\textsuperscript{$\infty,\heartsuit$} 
\ \quad 
Nancy F. Chen\textsuperscript{$\diamondsuit,\heartsuit$}
}
\\
\textsuperscript{$\infty$}IRIT, University of Toulouse, France\\
\textsuperscript{$\diamondsuit$}Agency for Science, Technology and Research (A*STAR), Singapore\\
\textsuperscript{$\heartsuit$}CNRS@CREATE, Singapore
\quad
\textsuperscript{$\S$}CNRS, IRIT, France\\
\texttt{galann.pennec@cnrsatcreate.sg},
\ \texttt{\{liu\_zhengyuan,nancy\_chen\}@a-star.edu.sg}\\
\ \texttt{\{nicholas.asher,philippe.muller\}@irit.fr}
}

\begin{document}

\newcommand{\cit}{{\color{red} cite }}

\newcommand{\hlred}[1]{{\sethlcolor{red}\hl{#1}}}
\newcommand{\hly}[1]{{\sethlcolor{yellow}\hl{#1}}}
\newcommand{\hlgr}[1]{{\sethlcolor{green}\hl{#1}}}
\newcommand{\boxred}[1]{{\fcolorbox{red}{white}{#1}}}

\maketitle
\begin{abstract}

Vision-Language Models (VLMs) are able to process increasingly longer videos. Yet, important visual information is easily lost throughout the entire context and missed by VLMs. Also, it is important to design tools that enable cost-effective analysis of lengthy video content. In this paper, we propose a clip selection method that targets key video moments to be included in a multimodal summary. We divide the video into short clips and generate compact visual descriptions of each using a lightweight video captioning model. These are then passed to a large language model (LLM), which selects the~$K$ clips containing the most relevant visual information for a multimodal summary. We evaluate our approach on reference clips for the task, automatically derived from full human-annotated screenplays and summaries in the MovieSum dataset. We further show that these reference clips (less than 6\% of the movie) are sufficient to build a complete multimodal summary of the movies in MovieSum. Using our clip selection method, we achieve a summarization performance close to that of these reference clips while capturing substantially more relevant video information than random clip selection. Importantly, we maintain low computational cost by relying on a lightweight captioning model.

\end{abstract}

\begin{figure*}[hbt]
\begin{center}
\includegraphics[width=0.88\textwidth]{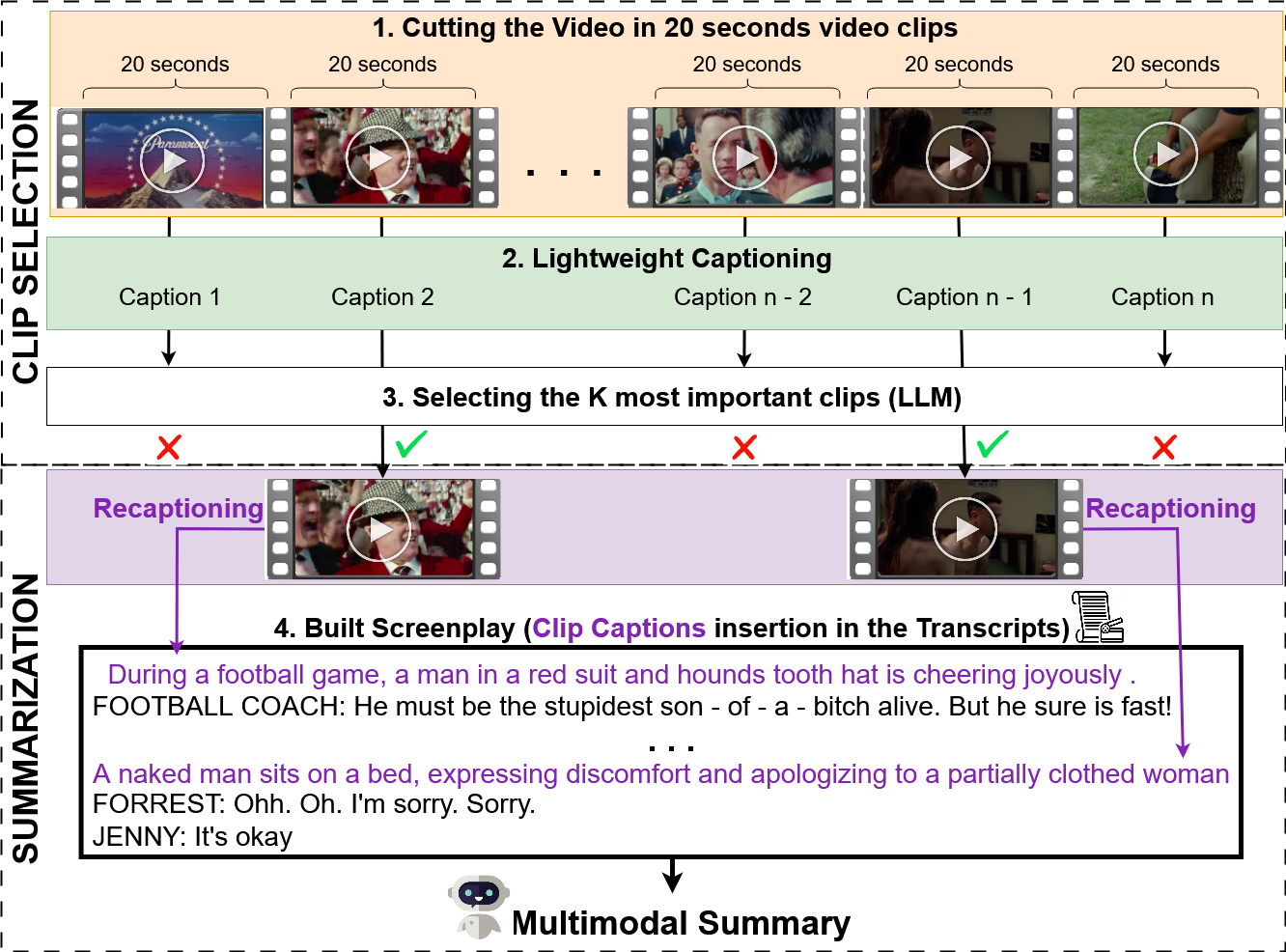}
\end{center}
\caption{\textbf{Our Clip Selection followed by Summarization.} 1) We segment the video into 20-second clips and generate lightweight captions for each. We then feed all the clip captions to an LLM to identify the top~$K$ clips that contain visually important information. 2) For summarization, we build a screenplay-like document by inserting the captions of the selected clips into the transcripts at the correct timestamps. We finally summarize these screenplays.}
\label{fig:pipeline}
\vspace{-0.3cm}
\end{figure*}

\section{Introduction}

Vision-Language Models (VLMs)~\citep{DBLP:journals/corr/abs-2502-13923, DBLP:journals/corr/abs-2503-12559, GPT-4o} have demonstrated improved capabilities in processing longer videos, particularly due to efficient pretraining~\citep{DBLP:conf/eccv/LiWJ24, DBLP:conf/eccv/WengHHCZ24,
DBLP:journals/corr/abs-2408-10188, DBLP:journals/corr/abs-2406-16852, DBLP:journals/corr/abs-2502-05173}.

However, performing inference on hour-long videos is costly and questions remain about how effectively VLMs handle longer contexts~\citep{DBLP:journals/corr/abs-2405-21075, DBLP:journals/corr/abs-2406-08035, DBLP:journals/corr/abs-2406-04264, DBLP:conf/nips/MangalamAM23}.
Notably, important visual elements are sometimes lost throughout the video, often causing VLMs to neglect or completely omit crucial information~\citep{DBLP:journals/corr/abs-2505-06594, DBLP:journals/corr/abs-2403-05262, DBLP:journals/corr/abs-2401-09774, DBLP:conf/cvpr/ShenYWYEW22, DBLP:journals/corr/abs-2408-12763}.

By observing that not all information in a video is relevant to a task, some strategies maintain a memory over past visual information when processing longer videos~\citep{DBLP:conf/cvpr/SongCWZZWCG0ZLH24, DBLP:conf/nips/QianDZZDLW24, DBLP:conf/cvpr/0004LJJCSSL24, DBLP:conf/icml/BalazevicSP0KH24, DBLP:journals/corr/abs-2403-14622}.
Similarly, in Long Video Understanding, the answer to a question about a video is usually contained within a small subset of key frames retrieved by video content selection methods~\citep{DBLP:journals/corr/abs-2406-09396, DBLP:conf/eccv/WangZZY24, DBLP:conf/nips/NarasimhanRD21}.

To the best of our knowledge, most of the above video content selection approaches have been designed for the vision modality alone with a limited focus on multimodal data where different modalities often overlap. Also, video content selection has been widely studied for Long Video Question Answering (LVQA) leaving Multimodal Video Summarization underexplored~\citep{DBLP:journals/corr/abs-2505-06594}.

In this paper, we make the observation that videos are often highly redundant across modalities, for instance, when what is shown visually is already conveyed through the dialogue or transcripts. We therefore consider the task of visually salient clip selection, meaning that we extract all clips containing relevant visual information that cannot be inferred from the transcripts alone.

We propose a cost-effective clip selection method (as shown in Figure~\ref{fig:pipeline}) and apply it to multimodal summarization of long videos such as movies from MovieSum~\citep{DBLP:conf/acl/SaxenaK24}\footnote{\url{https://huggingface.co/datasets/rohitsaxena/MovieSum}} which offer a reliable testbed due to their rich narratives, diverse multimodal cues, and their need for cross-modal integration.

Unlike LVQA that can assign confidence scores to individual frames or small frame sets, video summarization needs an understanding of the full context to identify key moments. To preserve temporal information, we treat clips, rather than individual frames, as the basic unit~\citep{DBLP:journals/corr/abs-2504-04471}. Moreover, instead of formulating the task as a binary classification over frames, we define it as selecting the top~$K$ most relevant clips from the entire video.

In Figure~\ref{fig:pipeline}, we divide the video into 20-second clips and generate a caption for each of them using a lightweight captioning model. The resulting captions are then passed to a LLM, which selects the top~$K$ most important clips to form the basis of the final multimodal summary.

Our contributions are as follows:

\begin{itemize}
    \item We propose visually salient clip selection as the task of retrieving all the video clips containing visual information relevant for a multimodal summary that cannot be inferred from the dialogue transcripts alone. 
    \item We introduce a lightweight clip selection strategy~(Figure~\ref{fig:pipeline}) allowing us to retrieve and target key video moments to generate long video summaries at a lower cost. We evaluate our approach based on reference clips for the task that we infer from MovieSum annotations.
    \item Using our clip selection strategy, we generate multimodal summaries of entire movies in MovieSum. Our summaries closely match those generated from the reference clips, while retrieving significantly more relevant visual information than random clip selection.
\end{itemize}

\section{Related Work}

\paragraph{From Text to Multimodal Summarization}

Summarization has a long history in the text modality, evolving from early rule-based approaches to neural sequence-to-sequence models~\cite{see2017get,liu-2021-coreference,chen-yang-2020-multi} and large language models~\cite{wang-etal-2023-instructive,tang-etal-2023-context}. These methods have been applied to diverse domains ranging from news and dialogues to scientific articles~\cite{feng2021survey,liu-2021-controllable,yasunaga2019scisummnet}, while addressing challenges such as salient span selection~\cite{kedzie-etal-2018-content,liu-2019-topic}, factual consistency~\cite{nan-etal-2021-entity,kryscinski2020evaluating}, and long-context understanding~\cite{gehrmann-etal-2018-bottom,kryscinski-etal-2022-booksum,liu2022dynamic}. With the development of multimodal approaches, the field has expanded beyond text to incorporate other modalities~\cite{li-etal-2017-multi}. Recent work explores multimodal summarization, integrating textual, audio, speech, and video data to generate feature-rich summaries ~\citep{DBLP:conf/eacl/PapalampidiL23,DBLP:journals/corr/abs-2505-06594}.

\paragraph{Video Content Selection}

Identifying important content from long videos has been addressed mostly in LVQA. Most of the time, the question to answer is used to query and retrieve relevant information (usually frames) from the whole video, whether in a zero-shot setting~\citep{DBLP:journals/corr/abs-2504-17447, DBLP:journals/corr/abs-2406-09396, DBLP:journals/corr/abs-2405-19209}, through pretraining~\citep{DBLP:conf/nips/Yu0YB23, DBLP:conf/iclr/YuJWCJZXSZWZS25, DBLP:conf/eccv/KorbarXTZT24} or via agentic approaches~\citep{DBLP:conf/eccv/WangZZY24, DBLP:journals/corr/abs-2412-10471, DBLP:journals/corr/abs-2504-04471}.

\paragraph{Efficient Long Video Summarization}
Although a summary can take the form of a video, such as a TV show recap or movie trailer~\citep{DBLP:conf/cvpr/SinghST24, DBLP:conf/aaai/PapalampidiKL21, DBLP:conf/cikm/ChenZZ24}, the present work generates long video summaries in text form instead.
Existing approaches to the task uniformly sample frames or clips throughout the original video either at a fixed rate~\citep{DBLP:conf/acl/0001WYMSZQLD25, DBLP:journals/kbs/AtriPGC21} or aligned with the scenes or dialogue utterances~\citep{DBLP:conf/acl/MahonL24, DBLP:conf/eacl/PapalampidiL23}. This uniform sampling results in inefficient video context management and VLMs easily missing out on valuable information by treating all video moments as equally important.
Noticing the variability in the importance of video moments, some approaches adopt alternative clip selection strategies for the video-to-text summarization task. For instance,~\citet{DBLP:journals/corr/abs-2505-06594} retrieves all video clips without any dialogue, arguing that they correlate with key visual moments of a movie or TV show.

In this work, we propose a clip selection method that identifies in zero-shot the top~$K$ visually salient clips containing important information to include in a multimodal video summary, and study its impact on the end summary.

\section{Clip Selection for Multimodal Video Summarization}
\label{sec:summ_model}
We approach multimodal summarization in two steps, treating clip selection as an intermediate task for summary generation. The complete pipeline is presented in Figure~\ref{fig:pipeline} and detailed in Section~\ref{subsec:the_pipeline}. In Section~\ref{subsec:pseudo_clips}, we further explain how clip selection is evaluated using the gold screenplay and groundtruth summary of a movie.

\subsection{Pipeline Overview}
\label{subsec:the_pipeline}

As shown in Figure~\ref{fig:pipeline}, we first segment the video into 20-second clips. Each clip is captioned using a lightweight VLM, and the resulting captions are all passed to an LLM for selection of the~$K$ clips containing important visual information. Clip selection is performed in either zero-shot or two-shot settings, depending on the prompts provided in Appendix~\ref{app:clip_selection}.

Following~\citet{DBLP:journals/corr/abs-2505-06594, DBLP:journals/corr/abs-2410-19809}, we build a screenplay-like document that efficiently represents the video's multimodal content, by combining the dialogue transcripts together with visual descriptions, for later summarization. To do so, we recaption the selected clips using a second, more robust VLM and insert them into the transcripts at the proper timestamp. As in~\citep{DBLP:conf/acl/MahonL24}, we could infer the timestamp of the transcripts utterances by aligning them with the corresponding audio in the video.

We finally summarize these screenplays using a customized prompt that incites the LLM to focus on multimodal cues from both the video captions and dialogue (see Appendix~\ref{subsec:summarization_prompt} for prompt details). We also place the marker~\texttt{`Caption:'} at the beginning of every clip caption in the screenplay to further facilitate the identification of important video content by the LLM.

\subsection{Clip Selection Reference}
\label{subsec:pseudo_clips}
Given the human-written screenplay and a reference summary of a movie we can extract all clips containing important visual information for a good multimodal understanding. Those clips serve as a reference for the task of clip selection in our evaluation~(section~\ref{subsec:clip_selection_metrics}). We proceed in three steps as follows. The first two steps are performed by an LLM in zero-shot, given the prompts in Appendix~\ref{subsec:reference_clips}.

\textbf{Step 1: Fact Identification} We decompose the groundtruth summary into a list of all its facts, each fact conveying a single piece of information (roughly equivalent to a simple clause).

\textbf{Step 2: Visual Fact Classification} We classify each groundtruth summary fact as~\texttt{Visual} (referring to the video) or~\texttt{Textual} (referring to the dialogue). For each fact, we ask the LLM to retrieve the information from within the human-written screenplay by specifically quoting the line. If the information comes from a clip caption in the screenplay, the fact is considered as~\texttt{Visual}. If it comes from the dialogue between the characters, we instead classify the fact as~\texttt{Textual}.

\textbf{Step 3: Reference Clips} For every~\texttt{Visual} fact in~\textbf{Step 2}, we locate the video segment that visually conveys this information. Using the screenplay timestamps, we define the clip to begin at the utterance immediately preceding the caption containing the~\texttt{Visual} fact and to end at the utterance immediately following it.

\section{Experimental Setup}
\label{sec:experiments}

\subsection{Datasets} \label{subsec:datasets}

We conduct our experiments on MovieSum~\citep{DBLP:conf/acl/SaxenaK24}, a summarization dataset of~2200 movies between~1950 and~2023, with equal splits of~200 movies each for validation and testing. The films span a diverse range of genres (comedy, drama, thriller,~\ldots) and have an average runtime of two hours. It includes detailed summaries~(635 words on average) referencing both video and dialogue modalities as well as long human-written screenplays~(25K words on average). Structurally, these screenplays are documents that interweave the dialogue transcripts with corresponding visual descriptions. We report all our experiments on the test split for which we purchased the videos.

\subsection{Clip Selection Metrics} \label{subsec:clip_selection_metrics}

Similar to previous work~\citep{DBLP:conf/iccv/MiechZATLS19, DBLP:conf/iccv/KrishnaHRFN17, DBLP:conf/acl/LeiYBB20}, we evaluate clip selection performance using Recall@K. The Recall@K denotes the ratio of reference clips retrieved by a clip selection method when we fix the number of selected clips to~$K$.

A reference clip~$r$ is deemed retrieved if the Intersection-over-Reference (IoR) between $r$ and a predicted clip $p$ is greater than~$\tau$, where $\tau$ is a fixed threshold. We define the IoR score as follows.
\begin{equation*}
\small{\text{IoR}(p,r)= \frac{|p \cap r|}{|r|}}
\end{equation*}
where $|p \cap r|$ denotes the temporal intersection length between $p$ and $r$.

\subsection{Summarization Metrics}

We report the summarization performance on both traditional and task-specific metrics.

\paragraph{Traditional Metrics} We report ROUGE-1 (r1), ROUGE-2 (r2), and ROUGE-Lsum (rlsum) using the python-rouge package, as well as METEOR scores computed with the \texttt{meteor\_score} function from \texttt{nltk.translate}.

\paragraph{\textsc{MFactSum}} We evaluate multimodal performance using~\textsc{MFactSum} metric~\citep{DBLP:journals/corr/abs-2505-06594}, which measures how effectively a multimodal summary captures the relevant information from both the video and dialogue. The metric computes two components: visual fact recall, assessing visual understanding, and textual fact recall, assessing textual understanding. The final multimodal score,~\textsc{MFactSum}, is obtained by averaging the two above components. Specifically, visual (resp. textual) fact recall refers here to the proportion of groundtruth summary facts originating from the video (resp. the dialogue) that are supported by the predicted summary.

\textsc{MFactSum} metrics rely on the following information: a decomposition of the groundtruth summary into facts, classified as~\texttt{Visual} or~\texttt{Textual}, and an assessment of whether these facts are supported by the predicted summary. 
The decomposition and classification can be done as described in section~\ref{subsec:pseudo_clips} by prompting an LLM in zero-shot. 
The final step of the evaluation uses the same LLM to judge if the predicted summary supports the~\texttt{Visual} and~\texttt{Textual} facts.
The prompt is given in Appendix~\ref{subsec:mfactsum_prompt}.

\begin{figure*}[htbp]
    \centering
    \includegraphics[width=0.3277\textwidth]{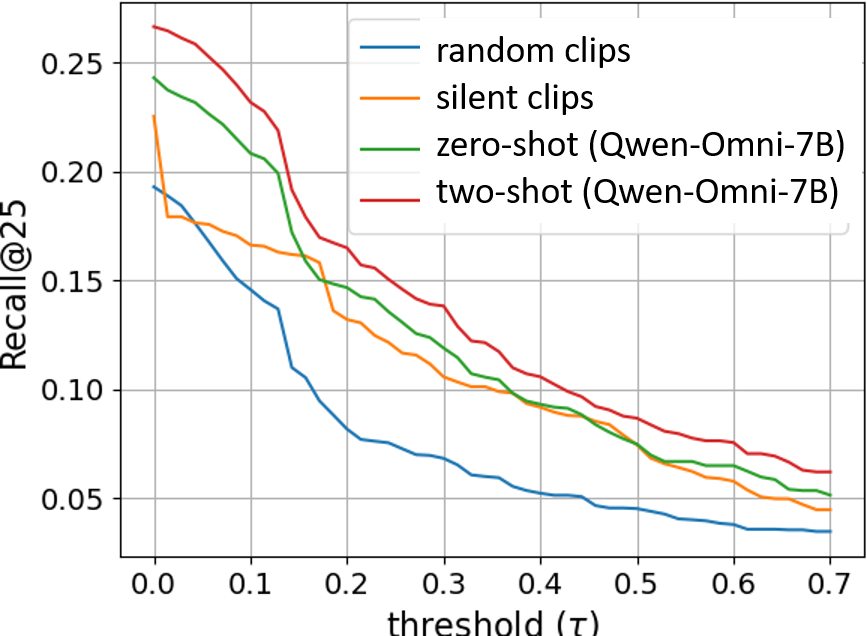}
    \includegraphics[width=0.3277\textwidth]{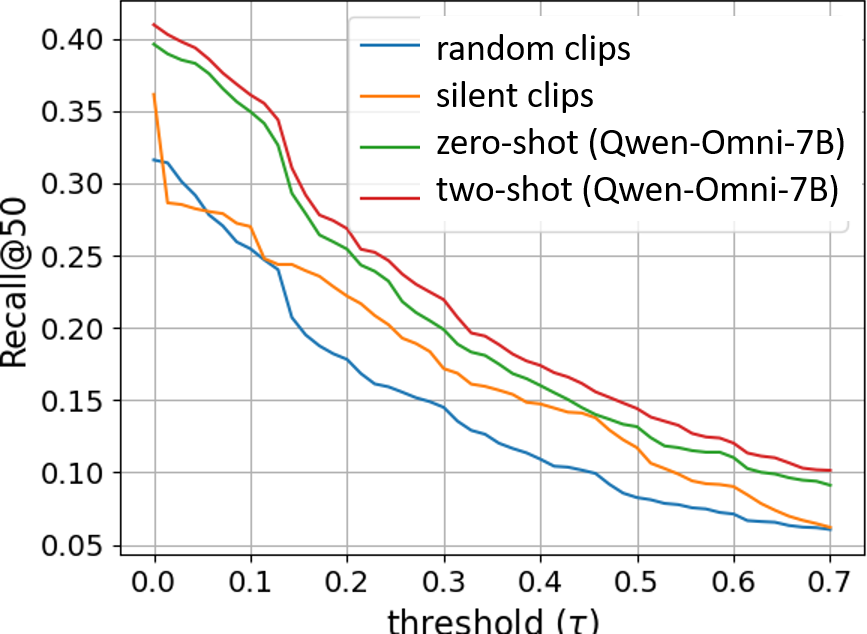}
    \includegraphics[width=0.3277\textwidth]{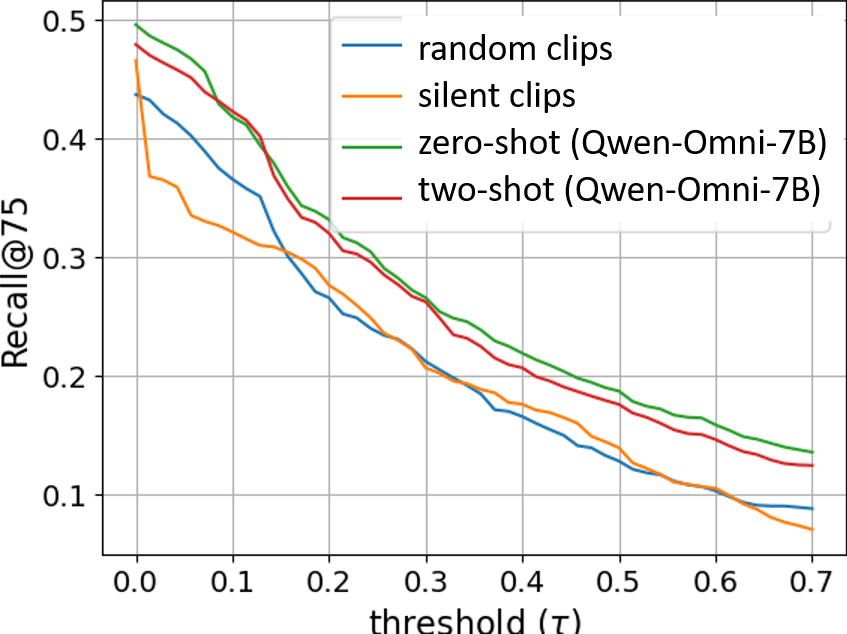}
    \caption{\textbf{Recall@K across varying thresholds~$\tau$.} Our clip selection outperforms other baselines regardless of the chosen threshold~$\tau$ used for the IoR matching.}
    \label{fig:clip_selection}
    \vspace{-0.3cm}
\end{figure*}

\subsection{Implementation Details}

We generate screenplay summaries for all the~200 movies from MovieSum test split using our pipeline in Figure~\ref{fig:pipeline}. We use either Qwen2.5-Omni-3B or Qwen2.5-Omni-7B as the lightweight captioning model, Gemini 2.5 Flash-Lite as the recaptioning model and Gemini 2.5 Flash for both the clip selection, summarization as well as for our evaluations with~\textsc{MFactSum}. While we choose Qwen2.5-Omni~\citep{DBLP:journals/corr/abs-2503-20215} for its high accuracy at a lower cost, Gemini 2.5 Flash-Lite~\citep{DBLP:journals/corr/abs-2507-06261} offers strong multimodal capabilities, making it well-suited for high quality recaptioning.
We also discuss results when replacing the summarization LLM by either Gemini 1.5 Flash~\citep{DBLP:journals/corr/abs-2403-05530} or Qwen2.5-72B-Instruct~\citep{DBLP:journals/corr/abs-2412-15115} in Appendix~\ref{app:experiments}. We disallow the thinking process and the use of external websites whenever using Gemini's API\footnote{\url{https://aistudio.google.com/}} in all our experiments. Because of the high API costs, results are presented for a single run only.

We always fix the target summary length to~1000 words in the prompt to all our baselines and models (see Appendix~\ref{subsec:summarization_prompt}). Also, we truncate the output summary to 1000 words for fair comparison between all settings. We do so because we are aware that some summarization metrics including the visual and textual recall as well as~\textsc{MFactSum} can increase mechanically with the summary length.

\section{Clip Selection Experiments}
\subsection{Baseline Methods} \label{sec:comparison_models}

In all our baselines, we fix~$K$ as the number of selected clips. We assign different values of~$K$ (25, 50, and 75) in practice, as shown in Figure~\ref{fig:clip_selection}.

\textbf{Random Clips} We randomly select~$K$ non-overlapping video clips of 20 seconds from the whole video.

\textbf{Silent Clips} All video clips that occur during a pause in the dialogue are considered~\citep{DBLP:journals/corr/abs-2505-06594}. Such clips are then sorted by decreasing duration and the~$K$ first are chosen. This heuristic baseline is motivated by the fact that silent scenes from a movie or TV show often highlight key visual moments and actions impacting the storyline.

\textbf{Our Clips} This corresponds to our clip selection method in either zero-shot or two-shot settings. In the main design~(Figure~\ref{fig:pipeline}), clip selection is performed by the LLM on captions generated by a lightweight captioning model (either Qwen2.5-Omni-3B or Qwen2.5-Omni-7B). For comparison, Table~\ref{tab:clip_selection} also reports results, instead, on the gold screenplay captions. In the latter case, the LLM is given all the captions present in the gold screenplay and is prompted to select the~$K$ most visually relevant ones using the same prompt as in the main setting~(Appendix~\ref{app:clip_selection}).

\begin{table}
\small
\center
\setlength{\tabcolsep}{2pt}
\resizebox{0.48\textwidth}{!}{
 \begin{tabular}{@{}lcccc|cc@{}}
\toprule & R@25 & R@50 & R@75 \\
\midrule
random clips & 6.83 & 14.50 & 21.22 \\
silent clips & 10.57 & 17.18 & 20.68 \\
\textit{ours zero-shot (Qwen2.5-Omni-3B)} & 11.14 & 18.34 & 25.40 \\
\textit{ours two-shot (Qwen2.5-Omni-3B)} & 10.82 & 19.85 & 25.63 \\
\textit{ours zero-shot (Qwen2.5-Omni-7B)} & 11.89 & 19.89 & \textbf{26.60} \\
\textit{ours two-shot (Qwen2.5-Omni-7B)} & \textbf{13.82} & \textbf{21.95} & 26.25 \\
\noalign{\vskip 0.5ex}
\cdashline{1-4}[2pt/2pt]
\noalign{\vskip 0.5ex}
\textit{ours zero-shot (gold screenplay captions)} & 39.56 & 51.22 & 66.79 \\
\textit{ours two-shot (gold screenplay captions)} & 39.79 & 55.04 & 67.33 \\
\bottomrule
\end{tabular}
}
\caption{\textbf{Evaluation of clip selection methods on the MovieSum test set.}
We report the Recall@K (R@K) for all studied clip selection strategies relative to the reference clips from section~\ref{subsec:pseudo_clips}. We also provide the scores when using the gold screenplays captions instead of Qwen2.5-Omni captions in our method. We fix the threshold~$\tau$ to~$0.3$ in the Recall@K computation. Note that there is on average~354 clips of~20 seconds in a movie from the MovieSum dataset. Therefore, $K=25$, $K=50$ and $K=75$ respectively corresponds, for our method, to retrieving about 7\%, 14\% and 28\% of the total movie duration length.} \label{tab:clip_selection}
\vspace{-0.3cm}
\end{table}

\begin{table*}
\center
\setlength{\tabcolsep}{2pt}
\resizebox{0.90\textwidth}{!}{
 \begin{tabular}{@{}p{1.5cm}<{\centering}p{1.2cm}<{\centering}p{1.2cm}<{\centering}p{1.2cm}<{\centering}|p{1.2cm}<{\centering}p{1.2cm}<{\centering}p{1.2cm}<{\centering}|p{2.0cm}<{\centering}@{}}
\toprule & vis-rec & text-rec & \textsc{MFS} & r1 & r2 & rlsum & METEOR \\
\midrule
\multicolumn{1}{l}{Transcripts (no clips)} & 14.42 & 26.89 & 20.65 & 44.66 & 10.35 & 42.64 & 32.12 \\
\multicolumn{1}{l}{Filtered Gold Screenplay (avg. 6\% clips)} & 32.84 & \textbf{35.63} & 34.23 & 45.73 & \textbf{13.63} & 43.90 & \textbf{36.24} \\
\multicolumn{1}{l}{Gold Screenplay (all clips)} & \textbf{34.47} & 35.48 & \textbf{34.97} & \textbf{47.43} & 11.88 & \textbf{45.34} & 34.06 \\

\bottomrule
\end{tabular}
}
\caption{\textbf{We only need about 6\% of the video information present in the gold screenplay to approximate a multimodal summary of the entire movie.} The filtered gold screenplay is obtained by keeping only the captions for the reference clips, accounting for about 6\% of all the captions. Evaluations are conducted on the MovieSum test set. We report the visual recall (vis-rec), textual recall (text-rec) and MFactSum denoted as MFS. We also include ROUGE-1 (r1), ROUGE-2 (r2), ROUGE-Lsum (rlsum) and METEOR. Best results are in \textbf{bold}.} \label{tab:motivation}
\end{table*}

\subsection{Results} \label{subsec:clip_results}

We report the Recall@K of various clip selection methods in both Figure~\ref{fig:clip_selection} and Table~\ref{tab:clip_selection}.
In Figure~\ref{fig:clip_selection}, our clip selection method (both zero-shot and two-shot) outperforms all tested baselines regardless of the chosen threshold~$\tau$ for the IoR matching. Also, using the two-shot examples further improves the performance of our method for lower values of~$K$~($K=25$ and $K=50$). For larger~$K$~($K=75$), the silent clip selection becomes less precise and performs close to the random selection baseline.

We also report the exact scores for different values of~$K$ when the threshold~$\tau$ is fixed to~$0.3$ in Table~\ref{tab:clip_selection}. The results highlight that the quality of the captions plays an important role for the task. Using the gold screenplay captions instead of Qwen2.5-Omni captions significantly boosts the Recall@K. Similarly, using a larger captioning model like Qwen2.5-Omni-7B instead of Qwen2.5-Omni-3B consistently improves the performance of our clip selection approach.

\subsection{Human Evaluation of the Clip Selection Reference} \label{sec:hhhhuman}

We validate the clip selection reference described in Section~\ref{subsec:pseudo_clips} through human evaluation conducted by two of the co-authors. The first annotator evaluated four randomly selected movies from the MovieSum test set: The Shining (1980), The Dark Knight (2008), The Imitation Game (2014), and Black Panther (2018). This evaluation covers 108 reference clips in total, providing a statistically meaningful sample for our analysis. We compute the agreement with the second annotator only on the movie The Dark Knight.

The human annotators are asked to watch each movie entirely and manually construct a human clip reference following the same procedure as in Section~\ref{subsec:pseudo_clips}. More precisely, given the groundtruth summary facts identified during~\textbf{Step 1}, the annotators retrieve all the video clips that support those facts~(\textbf{Steps 2 \& 3}). We find this step to have an accuracy of 84.6\% between our two annotators.

On the four movies, our clip selection reference achieves an F1 score of 86.5\% against the first human reference (see Appendix~\ref{app:human_eval_clip}).

\section{Multimodal Video Summarization Experiments}
\subsection{Baseline Methods}

We study the video summarization task under various settings, with results reported in Table~\ref{tab:main-results-MFactSum}.
For every setting, we always use the same exact summarization prompt defined in Appendix~\ref{subsec:summarization_prompt}.

\textbf{Transcripts (no video)}
We generate summaries from the transcripts alone.

\textbf{Built Screenplay~($K$ Clips)}
Following our pipeline~(Figure~\ref{fig:pipeline}), we build the screenplay from the~$K$ selected clips and generate a screenplay summary.
We also replace the clip selection component in our pipeline with alternative clip selection baselines from Section~\ref{sec:comparison_models} such as random clips or silent clips.

\textbf{Built Screenplay (reference clips)}
We build the screenplay from the reference clips we identified in Section~\ref{subsec:pseudo_clips}. We simply feed the reference clips directly into our summarization pipeline~(Figure~\ref{fig:pipeline}). This setting serves us as an upperbound as we inject the best possible clips into our pipeline.

\textbf{Gold Screenplay}
We generate summaries from the screenplay annotations given in MovieSum.

\begin{table*}
\center
\setlength{\tabcolsep}{2pt}
\resizebox{0.90\textwidth}{!}{
 \begin{tabular}{@{}p{1.5cm}<{\centering}p{1.3cm}<{\centering}p{1.3cm}<{\centering}p{1.3cm}<{\centering}|p{1.3cm}<{\centering}p{1.3cm}<{\centering}p{1.3cm}<{\centering}|p{1.8cm}<{\centering}@{}}
\toprule
& vis-rec & text-rec & \textsc{MFS} & r1 & r2 & rlsum & METEOR \\
\midrule
\multicolumn{1}{l}{\textbf{Transcripts (no video)}} & 14.42 & 26.89 & 20.65 & 44.66 & 10.35 & 42.64 & 32.12 \\
\noalign{\vskip 0.5ex}
\cdashline{1-8}[2pt/2pt]
\noalign{\vskip 0.5ex}
\multicolumn{1}{c}{\textbf{Built Screenplay (25 clips)}} & & & & & & & \\
\multicolumn{1}{l}{random clips} & 15.44 & 31.93 & 23.69 & 46.11 & 10.61 & 44.07 & 33.27 \\
\multicolumn{1}{l}{silent clips} & 16.33 & 32.61 & 24.47 & 46.24 & 10.95 & 44.11 & 33.24 \\
\multicolumn{1}{l}{\textit{our clips zero-shot (Qwen2.5-Omni-7B)}} & 20.81 & \textbf{35.64} & \textbf{28.23} & 46.29 & 11.20 & 44.16 & 33.59 \\
\multicolumn{1}{l}{\textit{our clips two-shot (Qwen2.5-Omni-7B)}} & \underline{21.05} & 34.01 & 27.53 & \textbf{46.90} & \textbf{11.55} & \textbf{44.69} & \textbf{33.93} \\
\noalign{\vskip 0.5ex}
\cdashline{1-8}[2pt/2pt]
\noalign{\vskip 0.5ex}
\multicolumn{1}{c}{\textbf{Built Screenplay (50 clips)}} & & & & & & & \\
\multicolumn{1}{l}{random clips} & 15.68 & 30.91 & 23.29 & 46.11 & 10.51 & 43.88 & 33.21 \\
\multicolumn{1}{l}{silent clips} & 17.53 & 31.82 & 24.67 & 46.10 & \underline{11.32} & 43.95 & 33.36 \\
\multicolumn{1}{l}{\textit{our clips zero-shot (Qwen2.5-Omni-7B)}} & 20.97 & 33.63 & 27.30 & \underline{46.54} & 11.22 & \underline{44.49} & \underline{33.63} \\
\multicolumn{1}{l}{\textit{our clips two-shot (Qwen2.5-Omni-7B)}} & 20.60 & \underline{34.22} & 27.41 & 46.22 & 10.97 & 44.10 & 33.46 \\
\noalign{\vskip 0.5ex}
\cdashline{1-8}[2pt/2pt]
\noalign{\vskip 0.5ex}
\multicolumn{1}{c}{\textbf{Built Screenplay (75 clips)}} & & & & & & & \\
\multicolumn{1}{l}{random clips} & 14.68 & 30.90 & 22.79 & 45.35 & 10.34 & 43.15 & 32.73 \\
\multicolumn{1}{l}{silent clips} & 19.43 & 32.42 & 25.92 & 46.00 & 10.84 & 43.83 & 33.37 \\
\multicolumn{1}{l}{\textit{our clips zero-shot (Qwen2.5-Omni-7B)}} & 20.87 & 31.25 & 26.06 & 46.04 & 10.75 & 43.87 & 33.13 \\
\multicolumn{1}{l}{\textit{our clips two-shot (Qwen2.5-Omni-7B)}} & \textbf{22.25} & 33.45 & \underline{27.85} & 46.53 & 10.80 & 44.32 & 33.55 \\
\noalign{\vskip 0.5ex}
\cdashline{1-8}[2pt/2pt]
\noalign{\vskip 0.5ex}
\multicolumn{1}{c}{\textbf{Built Screenplay (reference clips)}} & 22.43 & 35.38 & 28.90 & 47.28 & 11.67 & 45.21 & 34.14 \\
\midrule
\multicolumn{1}{l}{\textbf{Gold Screenplay}} & 34.47 & 35.48 & 34.97 & 47.43 & 11.88 & 45.34 & 34.06 \\

\bottomrule
\end{tabular}
}
\caption{\textbf{Summarization results on the MovieSum test set.} Except for the gold screenplay, all built screenplays in the Table are produced using Gemini 2.5 Flash-Lite as the recaptioning model. We always use Gemini 2.5 Flash as the summarization model.
Column descriptions are the same as in Table~\ref{tab:motivation}. Best results are in \textbf{bold}.} 
\label{tab:main-results-MFactSum}
\end{table*}

\subsection{Results} \label{subsec:summ_results}

\paragraph{MovieSum summaries are highly multimodal} We discover that a third of the summary content refers to video information. More precisely, we identify~35~\texttt{Visual} facts on average in MovieSum summaries.

\paragraph{6\% of the video is enough for a complete movie summary} Despite being highly multimodal, MovieSum summaries can be effectively built using a small fraction (6\%) of the gold screenplay captions. Specifically, using only the captions from the reference clips (21 clips on average) provides a summary nearly as informative as one built from the full screenplay~(Table~\ref{tab:motivation}). This finding strongly motivates the use of clip selection in long video summarization.

\paragraph{Our clip selection leads to summaries that better include multimodal information}
The summarization results in Table~\ref{tab:main-results-MFactSum} show that our clip selection method outperforms the other clip selection baselines especially on the visual recall, textual recall and~\textsc{MFactSum} metrics. In particular, we are able to retrieve substantially more relevant visual information (visual recall) than the random clip selection baseline. Remarkably, our performance is even close to that of the best possible clips (screenplay of the reference clips).
As noted by~\citep{DBLP:journals/corr/abs-2505-06594}, we found improvements to be less pronounced on traditional metrics such as ROUGE or METEOR as those metrics are not primarily designed for multimodality. All the above results were found to be similar when using other summarization models instead~(see Appendix~\ref{app:experiments}).

\textbf{Choice of~$K$ on the Summarization Performance}
Although the clip selection improves with larger values of~$K$ (see Recall@K in Table~\ref{tab:clip_selection}), this observation does not apply to the quality of the end summary. Indeed, the summarization performance reported in Table~\ref{tab:main-results-MFactSum} seems to saturate rather than monotonically increase with~$K$.

Since our summaries are constrained to a fixed target length (1000 words), we believe that growing values of~$K$ does not necessarily yield better summaries, as additional clips often exceed what the LLM can effectively leverage given the summary length constraint.

\textbf{Importance of the captioning quality}
The quality of the captions used to build the screenplay has a critical role. Indeed, summaries generated using Gemini 2.5 Flash-Lite for recaptioning capture significantly less visual information (visual recall) than those generated from the gold screenplay as input~(Table~\ref{tab:main-results-MFactSum}).

\begin{table*}
\center
\setlength{\tabcolsep}{2pt}
\small
\resizebox{0.88\textwidth}{!}{
 \begin{tabular}{@{}p{1.5cm}<{\centering}p{1.2cm}<{\centering}p{1.2cm}<{\centering}p{1.2cm}<{\centering}|p{1.2cm}<{\centering}p{1.2cm}<{\centering}p{1.2cm}<{\centering}|p{1.5cm}<{\centering}@{}}
\toprule & vis-rec & text-rec & \textsc{MFS} & r1 & r2 & rlsum & METEOR \\
\midrule
\multicolumn{1}{c}{\textbf{Built Screenplay (25 clips)}} & & & & & & & \\
\multicolumn{1}{l}{w/o recaptioning} & 19.31 & \textbf{36.71} & \textbf{28.01} & \textbf{47.07} & \textbf{11.64} & \textbf{44.80} & 33.79 \\
\multicolumn{1}{l}{with recaptioning (ours)} & \textbf{21.05} & 34.01 & 27.53 & 46.90 & 11.55 & 44.69 & \textbf{33.93} \\
\noalign{\vskip 0.5ex}
\cdashline{1-8}[2pt/2pt]
\noalign{\vskip 0.5ex}
\multicolumn{1}{c}{\textbf{Built Screenplay (50 clips)}} & & & & & & & \\
\multicolumn{1}{l}{w/o recaptioning} & 19.96 & 34.02 & 26.99 & 46.14 & \textbf{11.05} & 43.99 & 33.30 \\
\multicolumn{1}{l}{with recaptioning (ours)} & \textbf{20.60} & \textbf{34.22} & \textbf{27.41} & \textbf{46.22} & 10.97 & \textbf{44.10} & \textbf{33.46} \\
\noalign{\vskip 0.5ex}
\cdashline{1-8}[2pt/2pt]
\noalign{\vskip 0.5ex}
\multicolumn{1}{c}{\textbf{Built Screenplay (75 clips)}} & & & & & & & \\
\multicolumn{1}{l}{w/o recaptioning} & 21.15 & \textbf{34.73} & \textbf{27.94} & \textbf{46.79} & \textbf{11.58} & \textbf{44.71} & \textbf{33.66}  \\
\multicolumn{1}{l}{with recaptioning (ours)} & \textbf{22.25} & 33.45 & 27.85 & 46.53 & 10.80 & 44.32 & 33.55 \\

\bottomrule
\end{tabular}
}
\caption{\textbf{Effect of recaptioning on the summarization pipeline performance.} Recaptioning of visually significant moments with a stronger model (Gemini 2.5 Flash Lite) directly improves how well the generated summary captures important visual information (visual recall). In the above, we always perform clip selection using Qwen2.5-Omni-7B as the lightweight captioning model and summarization using Gemini 2.5 Flash.
Column descriptions are the same as in Table~\ref{tab:motivation}. Best results are in \textbf{bold}.} 
\label{tab:abl-recaptioning}
\vspace{-0.1cm}
\end{table*}

\begin{table}
\small
\center
\setlength{\tabcolsep}{2pt}
\resizebox{0.48\textwidth}{!}{
 \begin{tabular}{p{3.5cm}cccc|cc}
\toprule & R@25 & R@50 & R@75 \\
\midrule
\multicolumn{1}{l}{\textbf{our clips zero-shot}} & & & \\
w/o scene segmentation & \textbf{11.89} & \textbf{19.89} & \textbf{26.60} \\
with scene segmentation & 10.94 & 18.07 & 23.64 \\
\noalign{\vskip 0.5ex}
\cdashline{1-4}[2pt/2pt]
\noalign{\vskip 0.5ex}
\multicolumn{1}{l}{\textbf{our clips two-shot}} & & & \\
w/o scene segmentation & \textbf{13.82} & \textbf{21.95} & \textbf{26.25} \\
with scene segmentation & 10.36 & 18.40 & 24.98 \\
\bottomrule
\end{tabular}
}
\caption{\textbf{Effect of scene segmentation on our clip selection.} Scene segmentation does not positively impact the performance of our clip selection. In the above, we use Qwen2.5-Omni-7B as the lightweight captioning model. Column desciptions are the same as in Table~\ref{tab:clip_selection}. Best results are in \textbf{bold}.} \label{tab:abl_scene_segmentation}
\vspace{-0.3cm}
\end{table}

\section{Discussion} \label{subsec:discussion}

\paragraph{Ablation Study: Effect of Recaptioning on Summarization Quality}

Table~\ref{tab:abl-recaptioning} examines the impact of the recaptioning step on the overall summarization performance of our pipeline~(Figure~\ref{fig:pipeline}). In particular, we observe a consistent decrease in the visual recall when no recaptioning is being performed, indicating the importance of this step for capturing important visual information.

This finding reveals a clear division of labor between the two captioning stages in our pipeline. While lightweight captions are sufficient for identifying salient clips, they often miss finer visual details that are crucial for building accurate multimodal summaries. This design balances efficiency and accuracy: most of the video is processed cheaply, while the few clips that matter are described in depth to boost summary quality.

\paragraph{Comparing Video Segmentation Approaches for Clip Selection}

A natural alternative to our fixed 20-second clips in Section~\ref{subsec:the_pipeline} is scene segmentation, which divides the video into shorter scenes that better align with semantic shifts in the narrative. We infer these scenes in a zero-shot manner using the method from~\citep{DBLP:journals/corr/abs-2503-01201}.

On our test set, the average scene duration is 73 seconds. In order to match our original setup, we further subdivide each scene into shorter segments by uniformly splitting them so that the average segment is now of 20 seconds. This is to ensure that the two approaches are comparable while we still benefit from the scene boundaries given by the scene segmentation.

Despite being more natural, scene-based segmentation did not outperform our fixed 20-second clips (Table~\ref{tab:abl_scene_segmentation}). Since clip selection is done at the caption level, we believe that performance depends less on whether segment boundaries match with the scenes or are chosen arbitrarily.

\paragraph{Subjectivity of the Clip Selection and Summarization Tasks}
To evaluate the subjectivity of the clip selection task across different summary sources, we also collect summaries from The Movie Spoiler\footnote{\url{https://themoviespoiler.com/}}. From the 200 movies in our test set, we successfully retrieve 54 corresponding summaries. Following the same procedure described in~\ref{subsec:reference_clips}, we infer the clips for those summaries. The Movie Spoiler summaries are longer and we infer twice as many clips from those on average. We compute the overlap between the clips found in the two summary sources. The Movie Spoiler summaries recover about~48.8\% of the reference clips in MovieSum summaries. Also, when evaluated against the MovieSum reference, they achieved a visual recall of 66.2\% and textual recall of 61.1\%.

\section{Conclusion}

This paper tackles the dual challenges of long video summarization: the high computational cost and the risk of missing crucial visual information. We propose a cost-effective, clip selection for the task. Our method performs initial captioning of short video segments at a lower cost followed by selection of key visual moments by an LLM for inclusion into the multimodal summary.
Our experiments on the MovieSum dataset demonstrated that a small fraction of the movies, about~6\% of their content, is sufficient to generate a comprehensive multimodal summary, validating the core principle of our approach. Second, our proposed clip selection method significantly outperforms the tested baselines, capturing substantially more relevant visual information than random clip selection. Crucially, the summaries built from our selected clips achieve a performance close to those generated from a perfect set of reference clips, demonstrating the robustness of our selection strategy.

Future work could extend this methodology to other multimodal generative tasks and domains, and explore different selection criteria.
Overall, our findings suggest that focusing on minimal yet highly salient clips offers an efficient paradigm for understanding long-form video content.

\section*{Limitations}

The performance of clip selection is closely tied to the quality of the lightweight captioning~(Section~\ref{subsec:clip_results}), suggesting that improvements in smaller VLMs could yield further gains.

Adaptive clip selection strategies that dynamically choose~$K$ based on the video duration and density would be useful to explore. In the meantime, our experiments reveal the limited impact of varying~$K$ on the end summary and this is mainly due to the fixed length of the generated summary. Such adaptive strategies for varying~$K$ could be particularly beneficial in an unconstrained summarization setting, where the summary length is not fixed and this could be investigated in future work.

While our method outperforms the random clip selection baseline, it still incurs a computational cost, both in generating captions and choosing the~$K$ best clips. This cost is still lower than processing videos end-to-end using a high-performing VLM such as Gemini.

\section*{Acknowledgments}
We thank the anonymous reviewers for their feedback. This research is supported by the National Research Foundation, Prime Minister’s Office, Singapore under its Campus for Research Excellence and Technological Enterprise (CREATE) programme. Any opinions, findings and conclusions or recommendations expressed in this material are those of the authors and do not reflect the views of the National Research Foundation, Singapore.

\bibliography{custom}

@article{DBLP:journals/corr/abs-2403-05262,
  author       = {Yifan Zhang and
                  Weichen Yu and
                  Qingsong Wen and
                  Xue Wang and
                  Zhang Zhang and
                  Liang Wang and
                  Rong Jin and
                  Tieniu Tan},
  title        = {Debiasing Multimodal Large Language Models},
  journal      = {CoRR},
  volume       = {abs/2403.05262},
  year         = {2024},
  url          = {https://doi.org/10.48550/arXiv.2403.05262},
  doi          = {10.48550/ARXIV.2403.05262},
  eprinttype    = {arXiv},
  eprint       = {2403.05262},
  bibsource    = {dblp computer science bibliography, https://dblp.org}
}

@article{DBLP:journals/corr/abs-2401-09774,
  author       = {Taichi Nishimura and
                  Shota Nakada and
                  Masayoshi Kondo},
  title        = {On the Audio Hallucinations in Large Audio-Video Language Models},
  journal      = {CoRR},
  volume       = {abs/2401.09774},
  year         = {2024},
  url          = {https://doi.org/10.48550/arXiv.2401.09774},
  doi          = {10.48550/ARXIV.2401.09774},
  eprinttype    = {arXiv},
  eprint       = {2401.09774},
  bibsource    = {dblp computer science bibliography, https://dblp.org}
}

@inproceedings{kryscinski2020evaluating,
  title={Evaluating the factual consistency of abstractive text summarization},
  author={Kry{\'s}ci{\'n}ski, Wojciech and McCann, Bryan and Xiong, Caiming and Socher, Richard},
  booktitle={Proceedings of the 2020 conference on empirical methods in natural language processing (EMNLP)},
  pages={9332--9346},
  year={2020}
}

@inproceedings{gehrmann-etal-2018-bottom,
    title = "Bottom-Up Abstractive Summarization",
    author = "Gehrmann, Sebastian  and
      Deng, Yuntian  and
      Rush, Alexander",
    editor = "Riloff, Ellen  and
      Chiang, David  and
      Hockenmaier, Julia  and
      Tsujii, Jun{'}ichi",
    booktitle = "Proceedings of the 2018 Conference on Empirical Methods in Natural Language Processing",
    month = oct # "-" # nov,
    year = "2018",
    address = "Brussels, Belgium",
    publisher = "Association for Computational Linguistics",
    url = "https://aclanthology.org/D18-1443/",
    doi = "10.18653/v1/D18-1443",
    pages = "4098--4109",
}

@inproceedings{nan-etal-2021-entity,
    title = "Entity-level Factual Consistency of Abstractive Text Summarization",
    author = "Nan, Feng  and
      Nallapati, Ramesh  and
      Wang, Zhiguo  and
      Nogueira dos Santos, Cicero  and
      Zhu, Henghui  and
      Zhang, Dejiao  and
      McKeown, Kathleen  and
      Xiang, Bing",
    editor = "Merlo, Paola  and
      Tiedemann, Jorg  and
      Tsarfaty, Reut",
    booktitle = "Proceedings of the 16th Conference of the European Chapter of the Association for Computational Linguistics: Main Volume",
    month = apr,
    year = "2021",
    address = "Online",
    publisher = "Association for Computational Linguistics",
    url = "https://aclanthology.org/2021.eacl-main.235/",
    doi = "10.18653/v1/2021.eacl-main.235",
    pages = "2727--2733",
}

@inproceedings{DBLP:conf/eacl/PapalampidiL23,
  author       = {Pinelopi Papalampidi and
                  Mirella Lapata},
  editor       = {Andreas Vlachos and
                  Isabelle Augenstein},
  title        = {Hierarchical3{D} Adapters for Long Video-to-text Summarization},
  booktitle    = {Findings of the Association for Computational Linguistics: {EACL}
                  2023, Dubrovnik, Croatia, May 2-6, 2023},
  pages        = {1267--1290},
  publisher    = {Association for Computational Linguistics},
  year         = {2023},
  url          = {https://doi.org/10.18653/v1/2023.findings-eacl.96},
  doi          = {10.18653/V1/2023.FINDINGS-EACL.96},
  bibsource    = {dblp computer science bibliography, https://dblp.org}
}

@article{DBLP:journals/corr/abs-2410-19809,
  author       = {Louis Mahon and
                  Mirella Lapata},
  title        = {ScreenWriter: Automatic Screenplay Generation and Movie Summarisation},
  journal      = {CoRR},
  volume       = {abs/2410.19809},
  year         = {2024},
  url          = {https://doi.org/10.48550/arXiv.2410.19809},
  doi          = {10.48550/ARXIV.2410.19809},
  eprinttype    = {arXiv},
  eprint       = {2410.19809},
  bibsource    = {dblp computer science bibliography, https://dblp.org}
}

@inproceedings{DBLP:conf/acl/MahonL24,
  author       = {Louis Mahon and
                  Mirella Lapata},
  editor       = {Lun{-}Wei Ku and
                  Andre Martins and
                  Vivek Srikumar},
  title        = {A Modular Approach for Multimodal Summarization of {TV} Shows},
  booktitle    = {Proceedings of the 62nd Annual Meeting of the Association for Computational
                  Linguistics (Volume 1: Long Papers), {ACL} 2024, Bangkok, Thailand,
                  August 11-16, 2024},
  pages        = {8272--8291},
  publisher    = {Association for Computational Linguistics},
  year         = {2024},
  url          = {https://doi.org/10.18653/v1/2024.acl-long.450},
  doi          = {10.18653/V1/2024.ACL-LONG.450},
  bibsource    = {dblp computer science bibliography, https://dblp.org}
}

@misc{GPT-4o,
 author       = {OpenAI},
  title        = {Hello {GPT}-4o},
  year         = 2024,
  note         = {Accessed: 2024-11-6},
  url          = {https://openai.com/index/hello-gpt-4o}
}

@inproceedings{DBLP:conf/acl/SaxenaK24,
  author       = {Rohit Saxena and
                  Frank Keller},
  editor       = {Lun{-}Wei Ku and
                  Andre Martins and
                  Vivek Srikumar},
  title        = {Movie{S}um: An Abstractive Summarization Dataset for Movie Screenplays},
  booktitle    = {Findings of the Association for Computational Linguistics, {ACL} 2024,
                  Bangkok, Thailand and virtual meeting, August 11-16, 2024},
  pages        = {4043--4050},
  publisher    = {Association for Computational Linguistics},
  year         = {2024},
  url          = {https://doi.org/10.18653/v1/2024.findings-acl.239},
  doi          = {10.18653/V1/2024.FINDINGS-ACL.239},
  bibsource    = {dblp computer science bibliography, https://dblp.org}
}

@inproceedings{DBLP:conf/cvpr/SinghST24,
  author       = {Aditya Kumar Singh and
                  Dhruv Srivastava and
                  Makarand Tapaswi},
  title        = {"Previously on..." from Recaps to Story Summarization},
  booktitle    = {{IEEE/CVF} Conference on Computer Vision and Pattern Recognition,
                  {CVPR} 2024, Seattle, WA, USA, June 16-22, 2024},
  pages        = {13635--13646},
  publisher    = {{IEEE}},
  year         = {2024},
  url          = {https://doi.org/10.1109/CVPR52733.2024.01294},
  doi          = {10.1109/CVPR52733.2024.01294},
  bibsource    = {dblp computer science bibliography, https://dblp.org}
}

@inproceedings{DBLP:conf/cvpr/ShenYWYEW22,
  author       = {Yuhan Shen and
                  Linjie Yang and
                  Longyin Wen and
                  Haichao Yu and
                  Ehsan Elhamifar and
                  Heng Wang},
  title        = {Exploring the Role of Audio in Video Captioning},
  booktitle    = {{IEEE/CVF} Conference on Computer Vision and Pattern Recognition,
                  {CVPR} 2024 - Workshops, Seattle, WA, USA, June 17-18, 2024},
  pages        = {2090--2100},
  publisher    = {{IEEE}},
  year         = {2024},
  url          = {https://doi.org/10.1109/CVPRW63382.2024.00214},
  doi          = {10.1109/CVPRW63382.2024.00214},
  bibsource    = {dblp computer science bibliography, https://dblp.org}
}

@inproceedings{DBLP:conf/cikm/ChenZZ24,
  author       = {Brian Y. Chen and
                  Xiangyuan Zhao and
                  Yingnan Zhu},
  editor       = {Edoardo Serra and
                  Francesca Spezzano},
  title        = {Personalized Video Summarization by Multimodal Video Understanding},
  booktitle    = {Proceedings of the 33rd {ACM} International Conference on Information
                  and Knowledge Management, {CIKM} 2024, Boise, ID, USA, October 21-25,
                  2024},
  pages        = {4382--4389},
  publisher    = {{ACM}},
  year         = {2024},
  url          = {https://doi.org/10.1145/3627673.3680011},
  doi          = {10.1145/3627673.3680011},
  bibsource    = {dblp computer science bibliography, https://dblp.org}
}

@inproceedings{DBLP:conf/aaai/PapalampidiKL21,
  author       = {Pinelopi Papalampidi and
                  Frank Keller and
                  Mirella Lapata},
  title        = {Movie Summarization via Sparse Graph Construction},
  booktitle    = {Thirty-Fifth {AAAI} Conference on Artificial Intelligence, {AAAI}
                  2021, Thirty-Third Conference on Innovative Applications of Artificial
                  Intelligence, {IAAI} 2021, The Eleventh Symposium on Educational Advances
                  in Artificial Intelligence, {EAAI} 2021, Virtual Event, February 2-9,
                  2021},
  pages        = {13631--13639},
  publisher    = {{AAAI} Press},
  year         = {2021},
  url          = {https://doi.org/10.1609/aaai.v35i15.17607},
  doi          = {10.1609/AAAI.V35I15.17607},
  bibsource    = {dblp computer science bibliography, https://dblp.org}
}

@article{DBLP:journals/corr/abs-2405-21075,
  author       = {Chaoyou Fu and
                  Yuhan Dai and
                  Yondong Luo and
                  Lei Li and
                  Shuhuai Ren and
                  Renrui Zhang and
                  Zihan Wang and
                  Chenyu Zhou and
                  Yunhang Shen and
                  Mengdan Zhang and
                  Peixian Chen and
                  Yanwei Li and
                  Shaohui Lin and
                  Sirui Zhao and
                  Ke Li and
                  Tong Xu and
                  Xiawu Zheng and
                  Enhong Chen and
                  Rongrong Ji and
                  Xing Sun},
  title        = {Video-{MME}: The First-Ever Comprehensive Evaluation Benchmark of Multi-modal
                  {LLM}s in Video Analysis},
  journal      = {CoRR},
  volume       = {abs/2405.21075},
  year         = {2024},
  url          = {https://doi.org/10.48550/arXiv.2405.21075},
  doi          = {10.48550/ARXIV.2405.21075},
  eprinttype    = {arXiv},
  eprint       = {2405.21075},
  bibsource    = {dblp computer science bibliography, https://dblp.org}
}

@article{DBLP:journals/corr/abs-2406-08035,
  author       = {Weihan Wang and
                  Zehai He and
                  Wenyi Hong and
                  Yean Cheng and
                  Xiaohan Zhang and
                  Ji Qi and
                  Shiyu Huang and
                  Bin Xu and
                  Yuxiao Dong and
                  Ming Ding and
                  Jie Tang},
  title        = {{LVB}ench: An Extreme Long Video Understanding Benchmark},
  journal      = {CoRR},
  volume       = {abs/2406.08035},
  year         = {2024},
  url          = {https://doi.org/10.48550/arXiv.2406.08035},
  doi          = {10.48550/ARXIV.2406.08035},
  eprinttype    = {arXiv},
  eprint       = {2406.08035},
  bibsource    = {dblp computer science bibliography, https://dblp.org}
}

@article{DBLP:journals/corr/abs-2406-04264,
  author       = {Junjie Zhou and
                  Yan Shu and
                  Bo Zhao and
                  Boya Wu and
                  Shitao Xiao and
                  Xi Yang and
                  Yongping Xiong and
                  Bo Zhang and
                  Tiejun Huang and
                  Zheng Liu},
  title        = {{MLVU:} {A} Comprehensive Benchmark for Multi-Task Long Video Understanding},
  journal      = {CoRR},
  volume       = {abs/2406.04264},
  year         = {2024},
  url          = {https://doi.org/10.48550/arXiv.2406.04264},
  doi          = {10.48550/ARXIV.2406.04264},
  eprinttype    = {arXiv},
  eprint       = {2406.04264},
  bibsource    = {dblp computer science bibliography, https://dblp.org}
}

@article{DBLP:journals/corr/abs-2406-16852,
  author       = {Peiyuan Zhang and
                  Kaichen Zhang and
                  Bo Li and
                  Guangtao Zeng and
                  Jingkang Yang and
                  Yuanhan Zhang and
                  Ziyue Wang and
                  Haoran Tan and
                  Chunyuan Li and
                  Ziwei Liu},
  title        = {Long Context Transfer from Language to Vision},
  journal      = {CoRR},
  volume       = {abs/2406.16852},
  year         = {2024},
  url          = {https://doi.org/10.48550/arXiv.2406.16852},
  doi          = {10.48550/ARXIV.2406.16852},
  eprinttype    = {arXiv},
  eprint       = {2406.16852},
  bibsource    = {dblp computer science bibliography, https://dblp.org}
}

@inproceedings{DBLP:conf/nips/Yu0YB23,
  author       = {Shoubin Yu and
                  Jaemin Cho and
                  Prateek Yadav and
                  Mohit Bansal},
  editor       = {Alice Oh and
                  Tristan Naumann and
                  Amir Globerson and
                  Kate Saenko and
                  Moritz Hardt and
                  Sergey Levine},
  title        = {Self-Chained Image-Language Model for Video Localization and Question
                  Answering},
  booktitle    = {Advances in Neural Information Processing Systems 36: Annual Conference
                  on Neural Information Processing Systems 2023, NeurIPS 2023, New Orleans,
                  LA, USA, December 10 - 16, 2023},
  year         = {2023},
  url          = {http://papers.nips.cc/paper\_files/paper/2023/hash/f22a9af8dbb348952b08bd58d4734b50-Abstract-Conference.html},
  bibsource    = {dblp computer science bibliography, https://dblp.org}
}

@article{DBLP:journals/corr/abs-2403-05530,
  author       = {Machel Reid and
                  Nikolay Savinov and
                  Denis Teplyashin and
                  Dmitry Lepikhin and
                  Timothy P. Lillicrap and
                  Jean{-}Baptiste Alayrac and
                  Radu Soricut and
                  Angeliki Lazaridou and
                  Orhan Firat and
                  Julian Schrittwieser and
                  Ioannis Antonoglou and
                  Rohan Anil and
                  Sebastian Borgeaud and
                  Andrew M. Dai and
                  Katie Millican and
                  Ethan Dyer and
                  Mia Glaese and
                  Thibault Sottiaux and
                  Benjamin Lee and
                  Fabio Viola and
                  Malcolm Reynolds and
                  Yuanzhong Xu and
                  James Molloy and
                  Jilin Chen and
                  Michael Isard and
                  Paul Barham and
                  Tom Hennigan and
                  Ross McIlroy and
                  Melvin Johnson and
                  Johan Schalkwyk and
                  Eli Collins and
                  Eliza Rutherford and
                  Erica Moreira and
                  Kareem Ayoub and
                  Megha Goel and
                  Clemens Meyer and
                  Gregory Thornton and
                  Zhen Yang and
                  Henryk Michalewski and
                  Zaheer Abbas and
                  Nathan Schucher and
                  Ankesh Anand and
                  Richard Ives and
                  James Keeling and
                  Karel Lenc and
                  Salem Haykal and
                  Siamak Shakeri and
                  Pranav Shyam and
                  Aakanksha Chowdhery and
                  Roman Ring and
                  Stephen Spencer and
                  Eren Sezener and
                  et al.},
  title        = {Gemini 1.5: Unlocking multimodal understanding across millions of
                  tokens of context},
  journal      = {CoRR},
  volume       = {abs/2403.05530},
  year         = {2024},
  url          = {https://doi.org/10.48550/arXiv.2403.05530},
  doi          = {10.48550/ARXIV.2403.05530},
  eprinttype    = {arXiv},
  eprint       = {2403.05530},
  bibsource    = {dblp computer science bibliography, https://dblp.org}
}

@article{DBLP:journals/corr/abs-2408-12763,
  author       = {Jean Park and
                  Kuk Jin Jang and
                  Basam Alasaly and
                  Sriharsha Mopidevi and
                  Andrew Zolensky and
                  Eric Eaton and
                  Insup Lee and
                  Kevin Johnson},
  title        = {Assessing Modality Bias in Video Question Answering Benchmarks with
                  Multimodal Large Language Models},
  journal      = {CoRR},
  volume       = {abs/2408.12763},
  year         = {2024},
  url          = {https://doi.org/10.48550/arXiv.2408.12763},
  doi          = {10.48550/ARXIV.2408.12763},
  eprinttype    = {arXiv},
  eprint       = {2408.12763},
  bibsource    = {dblp computer science bibliography, https://dblp.org}
}

@inproceedings{DBLP:conf/acl/LeiYBB20,
  author       = {Jie Lei and
                  Licheng Yu and
                  Tamara L. Berg and
                  Mohit Bansal},
  editor       = {Dan Jurafsky and
                  Joyce Chai and
                  Natalie Schluter and
                  Joel R. Tetreault},
  title        = {{TVQA+:} Spatio-Temporal Grounding for Video Question Answering},
  booktitle    = {Proceedings of the 58th Annual Meeting of the Association for Computational
                  Linguistics, {ACL} 2020, Online, July 5-10, 2020},
  pages        = {8211--8225},
  publisher    = {Association for Computational Linguistics},
  year         = {2020},
  url          = {https://doi.org/10.18653/v1/2020.acl-main.730},
  doi          = {10.18653/V1/2020.ACL-MAIN.730},
  bibsource    = {dblp computer science bibliography, https://dblp.org}
}

@inproceedings{DBLP:conf/acl/0001WYMSZQLD25,
  author       = {Dongqi Liu and
                  Chenxi Whitehouse and
                  Xi Yu and
                  Louis Mahon and
                  Rohit Saxena and
                  Zheng Zhao and
                  Yifu Qiu and
                  Mirella Lapata and
                  Vera Demberg},
  editor       = {Wanxiang Che and
                  Joyce Nabende and
                  Ekaterina Shutova and
                  Mohammad Taher Pilehvar},
  title        = {What Is That Talk About? {A} Video-to-Text Summarization Dataset for
                  Scientific Presentations},
  booktitle    = {Proceedings of the 63rd Annual Meeting of the Association for Computational
                  Linguistics (Volume 1: Long Papers), {ACL} 2025, Vienna, Austria,
                  July 27 - August 1, 2025},
  pages        = {6187--6210},
  publisher    = {Association for Computational Linguistics},
  year         = {2025},
  url          = {https://aclanthology.org/2025.acl-long.310/},
  bibsource    = {dblp computer science bibliography, https://dblp.org}
}

@article{DBLP:journals/kbs/AtriPGC21,
  author       = {Yash Kumar Atri and
                  Shraman Pramanick and
                  Vikram Goyal and
                  Tanmoy Chakraborty},
  title        = {See, hear, read: Leveraging multimodality with guided attention for
                  abstractive text summarization},
  journal      = {Knowl. Based Syst.},
  volume       = {227},
  pages        = {107152},
  year         = {2021},
  url          = {https://doi.org/10.1016/j.knosys.2021.107152},
  doi          = {10.1016/J.KNOSYS.2021.107152},
  bibsource    = {dblp computer science bibliography, https://dblp.org}
}

@inproceedings{DBLP:conf/nips/MangalamAM23,
  author       = {Karttikeya Mangalam and
                  Raiymbek Akshulakov and
                  Jitendra Malik},
  editor       = {Alice Oh and
                  Tristan Naumann and
                  Amir Globerson and
                  Kate Saenko and
                  Moritz Hardt and
                  Sergey Levine},
  title        = {EgoSchema: {A} Diagnostic Benchmark for Very Long-form Video Language
                  Understanding},
  booktitle    = {Advances in Neural Information Processing Systems 36: Annual Conference
                  on Neural Information Processing Systems 2023, NeurIPS 2023, New Orleans,
                  LA, USA, December 10 - 16, 2023},
  year         = {2023},
  url          = {http://papers.nips.cc/paper\_files/paper/2023/hash/90ce332aff156b910b002ce4e6880dec-Abstract-Datasets\_and\_Benchmarks.html},
  bibsource    = {dblp computer science bibliography, https://dblp.org}
}

@article{DBLP:journals/corr/abs-2504-17447,
  author       = {De{-}An Huang and
                  Subhashree Radhakrishnan and
                  Zhiding Yu and
                  Jan Kautz},
  title        = {{FRAG:} Frame Selection Augmented Generation for Long Video and Long
                  Document Understanding},
  journal      = {CoRR},
  volume       = {abs/2504.17447},
  year         = {2025},
  url          = {https://doi.org/10.48550/arXiv.2504.17447},
  doi          = {10.48550/ARXIV.2504.17447},
  eprinttype    = {arXiv},
  eprint       = {2504.17447},
  bibsource    = {dblp computer science bibliography, https://dblp.org}
}

@article{DBLP:journals/corr/abs-2406-09396,
  author       = {Jongwoo Park and
                  Kanchana Ranasinghe and
                  Kumara Kahatapitiya and
                  Wonjeong Ryoo and
                  Donghyun Kim and
                  Michael S. Ryoo},
  title        = {Too Many Frames, not all Useful: Efficient Strategies for Long-Form
                  Video {QA}},
  journal      = {CoRR},
  volume       = {abs/2406.09396},
  year         = {2024},
  url          = {https://doi.org/10.48550/arXiv.2406.09396},
  doi          = {10.48550/ARXIV.2406.09396},
  eprinttype    = {arXiv},
  eprint       = {2406.09396},
  bibsource    = {dblp computer science bibliography, https://dblp.org}
}

@article{DBLP:journals/corr/abs-2405-19209,
  author       = {Ziyang Wang and
                  Shoubin Yu and
                  Elias Stengel{-}Eskin and
                  Jaehong Yoon and
                  Feng Cheng and
                  Gedas Bertasius and
                  Mohit Bansal},
  title        = {Video{T}ree: Adaptive Tree-based Video Representation for {LLM} Reasoning
                  on Long Videos},
  journal      = {CoRR},
  volume       = {abs/2405.19209},
  year         = {2024},
  url          = {https://doi.org/10.48550/arXiv.2405.19209},
  doi          = {10.48550/ARXIV.2405.19209},
  eprinttype    = {arXiv},
  eprint       = {2405.19209},
  bibsource    = {dblp computer science bibliography, https://dblp.org}
}

@inproceedings{DBLP:conf/eccv/WangZZY24,
  author       = {Xiaohan Wang and
                  Yuhui Zhang and
                  Orr Zohar and
                  Serena Yeung{-}Levy},
  editor       = {Ales Leonardis and
                  Elisa Ricci and
                  Stefan Roth and
                  Olga Russakovsky and
                  Torsten Sattler and
                  G{\"{u}}l Varol},
  title        = {Video{A}gent: Long-Form Video Understanding with Large Language Model
                  as Agent},
  booktitle    = {Computer Vision - {ECCV} 2024 - 18th European Conference, Milan, Italy,
                  September 29-October 4, 2024, Proceedings, Part {LXXX}},
  series       = {Lecture Notes in Computer Science},
  volume       = {15138},
  pages        = {58--76},
  publisher    = {Springer},
  year         = {2024},
  url          = {https://doi.org/10.1007/978-3-031-72989-8\_4},
  doi          = {10.1007/978-3-031-72989-8\_4},
  bibsource    = {dblp computer science bibliography, https://dblp.org}
}

@article{DBLP:journals/corr/abs-2412-10471,
  author       = {Zeyuan Yang and
                  Delin Chen and
                  Xueyang Yu and
                  Maohao Shen and
                  Chuang Gan},
  title        = {{VCA:} Video Curious Agent for Long Video Understanding},
  journal      = {CoRR},
  volume       = {abs/2412.10471},
  year         = {2024},
  url          = {https://doi.org/10.48550/arXiv.2412.10471},
  doi          = {10.48550/ARXIV.2412.10471},
  eprinttype    = {arXiv},
  eprint       = {2412.10471},
  bibsource    = {dblp computer science bibliography, https://dblp.org}
}

@article{DBLP:journals/corr/abs-2504-04471,
  author       = {Zhuo Zhi and
                  Qiangqiang Wu and
                  Minghe shen and
                  Wenbo Li and
                  Yinchuan Li and
                  Kun Shao and
                  Kaiwen Zhou},
  title        = {Video{A}gent2: Enhancing the {LLM}-Based Agent System for Long-Form Video
                  Understanding by Uncertainty-Aware {C}o{T}},
  journal      = {CoRR},
  volume       = {abs/2504.04471},
  year         = {2025},
  url          = {https://doi.org/10.48550/arXiv.2504.04471},
  doi          = {10.48550/ARXIV.2504.04471},
  eprinttype    = {arXiv},
  eprint       = {2504.04471},
  bibsource    = {dblp computer science bibliography, https://dblp.org}
}

@article{DBLP:journals/corr/abs-2503-12559,
  author       = {Xiao Wang and
                  Qingyi Si and
                  Jianlong Wu and
                  Shiyu Zhu and
                  Li Cao and
                  Liqiang Nie},
  title        = {{A}da{R}e{T}a{K}e: Adaptive Redundancy Reduction to Perceive Longer for Video-language
                  Understanding},
  journal      = {CoRR},
  volume       = {abs/2503.12559},
  year         = {2025},
  url          = {https://doi.org/10.48550/arXiv.2503.12559},
  doi          = {10.48550/ARXIV.2503.12559},
  eprinttype    = {arXiv},
  eprint       = {2503.12559},
  bibsource    = {dblp computer science bibliography, https://dblp.org}
}

@article{DBLP:journals/corr/abs-2502-13923,
  author       = {Shuai Bai and
                  Keqin Chen and
                  Xuejing Liu and
                  Jialin Wang and
                  Wenbin Ge and
                  Sibo Song and
                  Kai Dang and
                  Peng Wang and
                  Shijie Wang and
                  Jun Tang and
                  Humen Zhong and
                  Yuanzhi Zhu and
                  Ming{-}Hsuan Yang and
                  Zhaohai Li and
                  Jianqiang Wan and
                  Pengfei Wang and
                  Wei Ding and
                  Zheren Fu and
                  Yiheng Xu and
                  Jiabo Ye and
                  Xi Zhang and
                  Tianbao Xie and
                  Zesen Cheng and
                  Hang Zhang and
                  Zhibo Yang and
                  Haiyang Xu and
                  Junyang Lin},
  title        = {Qwen2.5-{VL} Technical Report},
  journal      = {CoRR},
  volume       = {abs/2502.13923},
  year         = {2025},
  url          = {https://doi.org/10.48550/arXiv.2502.13923},
  doi          = {10.48550/ARXIV.2502.13923},
  eprinttype    = {arXiv},
  eprint       = {2502.13923},
  bibsource    = {dblp computer science bibliography, https://dblp.org}
}

@inproceedings{see2017get,
  title={Get To The Point: Summarization with Pointer-Generator Networks},
  author={See, Abigail and Liu, Peter J and Manning, Christopher D},
  booktitle={Proceedings of the 55th Annual Meeting of the Association for Computational Linguistics (Volume 1: Long Papers)},
  pages={1073--1083},
  year={2017}
}

@inproceedings{liu-2021-controllable,
    title = "Controllable Neural Dialogue Summarization with Personal Named Entity Planning",
    author = "Liu, Zhengyuan  and
      Chen, Nancy",
    booktitle = "Proceedings of the 2021 Conference on Empirical Methods in Natural Language Processing",
    month = nov,
    year = "2021",
    address = "Online and Punta Cana, Dominican Republic",
    publisher = "Association for Computational Linguistics",
    url = "https://aclanthology.org/2021.emnlp-main.8/",
    doi = "10.18653/v1/2021.emnlp-main.8",
    pages = "92--106"
}

@article{feng2021survey,
  title={A survey on dialogue summarization: Recent advances and new frontiers},
  author={Feng, Xiachong and Feng, Xiaocheng and Qin, Bing},
  journal={arXiv preprint arXiv:2107.03175},
  year={2021}
}

@inproceedings{yasunaga2019scisummnet,
  title={Scisummnet: A large annotated corpus and content-impact models for scientific paper summarization with citation networks},
  author={Yasunaga, Michihiro and Kasai, Jungo and Zhang, Rui and Fabbri, Alexander R and Li, Irene and Friedman, Dan and Radev, Dragomir R},
  booktitle={Proceedings of the AAAI conference on artificial intelligence},
  volume={33},
  number={01},
  pages={7386--7393},
  year={2019}
}

@inproceedings{liu2022dynamic,
  title={Dynamic Sliding Window Modeling for Abstractive Meeting Summarization},
  author={Liu, Zhengyuan and Chen, Nancy},
  booktitle={Proc. Interspeech 2022},
  pages={5150--5154},
  year={2022}
}

@inproceedings{kedzie-etal-2018-content,
    title = "Content Selection in Deep Learning Models of Summarization",
    author = "Kedzie, Chris  and
      McKeown, Kathleen  and
      Daum{\'e} III, Hal",
    editor = "Riloff, Ellen  and
      Chiang, David  and
      Hockenmaier, Julia  and
      Tsujii, Jun{'}ichi",
    booktitle = "Proceedings of the 2018 Conference on Empirical Methods in Natural Language Processing",
    month = oct # "-" # nov,
    year = "2018",
    address = "Brussels, Belgium",
    publisher = "Association for Computational Linguistics",
    url = "https://aclanthology.org/D18-1208/",
    doi = "10.18653/v1/D18-1208",
    pages = "1818--1828",
}

@inproceedings{li-etal-2017-multi,
    title = "Multi-modal Summarization for Asynchronous Collection of Text, Image, Audio and Video",
    author = "Li, Haoran  and
      Zhu, Junnan  and
      Ma, Cong  and
      Zhang, Jiajun  and
      Zong, Chengqing",
    editor = "Palmer, Martha  and
      Hwa, Rebecca  and
      Riedel, Sebastian",
    booktitle = "Proceedings of the 2017 Conference on Empirical Methods in Natural Language Processing",
    month = sep,
    year = "2017",
    address = "Copenhagen, Denmark",
    publisher = "Association for Computational Linguistics",
    url = "https://aclanthology.org/D17-1114/",
    doi = "10.18653/v1/D17-1114",
    pages = "1092--1102",
}

@inproceedings{kryscinski-etal-2022-booksum,
    title = "{BOOKSUM}: A Collection of Datasets for Long-form Narrative Summarization",
    author = "Kryscinski, Wojciech  and
      Rajani, Nazneen  and
      Agarwal, Divyansh  and
      Xiong, Caiming  and
      Radev, Dragomir",
    editor = "Goldberg, Yoav  and
      Kozareva, Zornitsa  and
      Zhang, Yue",
    booktitle = "Findings of the Association for Computational Linguistics: EMNLP 2022",
    month = dec,
    year = "2022",
    address = "Abu Dhabi, United Arab Emirates",
    publisher = "Association for Computational Linguistics",
    url = "https://aclanthology.org/2022.findings-emnlp.488/",
    doi = "10.18653/v1/2022.findings-emnlp.488",
    pages = "6536--6558",
}

@inproceedings{chen-yang-2020-multi,
    title = "Multi-View Sequence-to-Sequence Models with Conversational Structure for Abstractive Dialogue Summarization",
    author = "Chen, Jiaao  and
      Yang, Diyi",
    editor = "Webber, Bonnie  and
      Cohn, Trevor  and
      He, Yulan  and
      Liu, Yang",
    booktitle = "Proceedings of the 2020 Conference on Empirical Methods in Natural Language Processing (EMNLP)",
    month = nov,
    year = "2020",
    address = "Online",
    publisher = "Association for Computational Linguistics",
    url = "https://aclanthology.org/2020.emnlp-main.336/",
    doi = "10.18653/v1/2020.emnlp-main.336",
    pages = "4106--4118",
}

@inproceedings{tang-etal-2023-context,
    title = "In-context Learning of Large Language Models for Controlled Dialogue Summarization: A Holistic Benchmark and Empirical Analysis",
    author = "Tang, Yuting  and
      Puduppully, Ratish  and
      Liu, Zhengyuan  and
      Chen, Nancy",
    editor = "Dong, Yue  and
      Xiao, Wen  and
      Wang, Lu  and
      Liu, Fei  and
      Carenini, Giuseppe",
    booktitle = "Proceedings of the 4th New Frontiers in Summarization Workshop",
    month = dec,
    year = "2023",
    address = "Singapore",
    publisher = "Association for Computational Linguistics",
    url = "https://aclanthology.org/2023.newsum-1.6/",
    doi = "10.18653/v1/2023.newsum-1.6",
    pages = "56--67",
}

@inproceedings{wang-etal-2023-instructive,
    title = "Instructive Dialogue Summarization with Query Aggregations",
    author = "Wang, Bin  and
      Liu, Zhengyuan  and
      Chen, Nancy",
    editor = "Bouamor, Houda  and
      Pino, Juan  and
      Bali, Kalika",
    booktitle = "Proceedings of the 2023 Conference on Empirical Methods in Natural Language Processing",
    month = dec,
    year = "2023",
    address = "Singapore",
    publisher = "Association for Computational Linguistics",
    url = "https://aclanthology.org/2023.emnlp-main.474/",
    doi = "10.18653/v1/2023.emnlp-main.474",
    pages = "7630--7653",
}

@inproceedings{liu-2019-topic,
  title={Topic-aware pointer-generator networks for summarizing spoken conversations},
  author={Liu, Zhengyuan and Ng, Angela and Lee, Sheldon and Aw, Ai Ti and Chen, Nancy F},
  booktitle={2019 IEEE Automatic Speech Recognition and Understanding Workshop (ASRU)},
  pages={814--821},
  year={2019},
  organization={IEEE}
}

@inproceedings{liu-2021-coreference,
    title = "Coreference-Aware Dialogue Summarization",
    author = "Liu, Zhengyuan  and
      Shi, Ke  and
      Chen, Nancy",
    booktitle = "Proceedings of the 22nd Annual Meeting of the Special Interest Group on Discourse and Dialogue",
    month = jul,
    year = "2021",
    address = "Singapore and Online",
    publisher = "Association for Computational Linguistics",
    url = "https://aclanthology.org/2021.sigdial-1.53/",
    doi = "10.18653/v1/2021.sigdial-1.53",
    pages = "509--519"
}

@article{DBLP:journals/corr/abs-2505-06594,
  author       = {Galann Pennec and
                  Zhengyuan Liu and
                  Nicholas Asher and
                  Philippe Muller and
                  Nancy F. Chen},
  title        = {Integrating Video and Text: {A} Balanced Approach to Multimodal Summary
                  Generation and Evaluation},
  journal      = {CoRR},
  volume       = {abs/2505.06594},
  year         = {2025},
  url          = {https://doi.org/10.48550/arXiv.2505.06594},
  doi          = {10.48550/ARXIV.2505.06594},
  eprinttype    = {arXiv},
  eprint       = {2505.06594},
  bibsource    = {dblp computer science bibliography, https://dblp.org}
}

@inproceedings{DBLP:conf/eccv/LiWJ24,
  author       = {Yanwei Li and
                  Chengyao Wang and
                  Jiaya Jia},
  editor       = {Ales Leonardis and
                  Elisa Ricci and
                  Stefan Roth and
                  Olga Russakovsky and
                  Torsten Sattler and
                  G{\"{u}}l Varol},
  title        = {{LL}a{MA}-{VID}: An Image is Worth 2 Tokens in Large Language Models},
  booktitle    = {Computer Vision - {ECCV} 2024 - 18th European Conference, Milan, Italy,
                  September 29-October 4, 2024, Proceedings, Part {XLVI}},
  series       = {Lecture Notes in Computer Science},
  volume       = {15104},
  pages        = {323--340},
  publisher    = {Springer},
  year         = {2024},
  url          = {https://doi.org/10.1007/978-3-031-72952-2\_19},
  doi          = {10.1007/978-3-031-72952-2\_19},
  bibsource    = {dblp computer science bibliography, https://dblp.org}
}

@inproceedings{DBLP:conf/cvpr/SongCWZZWCG0ZLH24,
  author       = {Enxin Song and
                  Wenhao Chai and
                  Guanhong Wang and
                  Yucheng Zhang and
                  Haoyang Zhou and
                  Feiyang Wu and
                  Haozhe Chi and
                  Xun Guo and
                  Tian Ye and
                  Yanting Zhang and
                  Yan Lu and
                  Jenq{-}Neng Hwang and
                  Gaoang Wang},
  title        = {Movie{C}hat: From Dense Token to Sparse Memory for Long Video Understanding},
  booktitle    = {{IEEE/CVF} Conference on Computer Vision and Pattern Recognition,
                  {CVPR} 2024, Seattle, WA, USA, June 16-22, 2024},
  pages        = {18221--18232},
  publisher    = {{IEEE}},
  year         = {2024},
  url          = {https://doi.org/10.1109/CVPR52733.2024.01725},
  doi          = {10.1109/CVPR52733.2024.01725},
  bibsource    = {dblp computer science bibliography, https://dblp.org}
}

@inproceedings{DBLP:conf/eccv/WengHHCZ24,
  author       = {Yuetian Weng and
                  Mingfei Han and
                  Haoyu He and
                  Xiaojun Chang and
                  Bohan Zhuang},
  editor       = {Ales Leonardis and
                  Elisa Ricci and
                  Stefan Roth and
                  Olga Russakovsky and
                  Torsten Sattler and
                  G{\"{u}}l Varol},
  title        = {Long{VLM}: Efficient Long Video Understanding via Large Language Models},
  booktitle    = {Computer Vision - {ECCV} 2024 - 18th European Conference, Milan, Italy,
                  September 29-October 4, 2024, Proceedings, Part {XXXIII}},
  series       = {Lecture Notes in Computer Science},
  volume       = {15091},
  pages        = {453--470},
  publisher    = {Springer},
  year         = {2024},
  url          = {https://doi.org/10.1007/978-3-031-73414-4\_26},
  doi          = {10.1007/978-3-031-73414-4\_26},
  bibsource    = {dblp computer science bibliography, https://dblp.org}
}

@inproceedings{DBLP:conf/nips/QianDZZDLW24,
  author       = {Rui Qian and
                  Xiaoyi Dong and
                  Pan Zhang and
                  Yuhang Zang and
                  Shuangrui Ding and
                  Dahua Lin and
                  Jiaqi Wang},
  editor       = {Amir Globersons and
                  Lester Mackey and
                  Danielle Belgrave and
                  Angela Fan and
                  Ulrich Paquet and
                  Jakub M. Tomczak and
                  Cheng Zhang},
  title        = {Streaming Long Video Understanding with Large Language Models},
  booktitle    = {Advances in Neural Information Processing Systems 38: Annual Conference
                  on Neural Information Processing Systems 2024, NeurIPS 2024, Vancouver,
                  BC, Canada, December 10 - 15, 2024},
  year         = {2024},
  url          = {http://papers.nips.cc/paper\_files/paper/2024/hash/d7ce06e9293c3d8e6cb3f80b4157f875-Abstract-Conference.html},
  bibsource    = {dblp computer science bibliography, https://dblp.org}
}

@article{DBLP:journals/corr/abs-2408-10188,
  author       = {Fuzhao Xue and
                  Yukang Chen and
                  Dacheng Li and
                  Qinghao Hu and
                  Ligeng Zhu and
                  Xiuyu Li and
                  Yunhao Fang and
                  Haotian Tang and
                  Shang Yang and
                  Zhijian Liu and
                  Ethan He and
                  Hongxu Yin and
                  Pavlo Molchanov and
                  Jan Kautz and
                  Linxi Fan and
                  Yuke Zhu and
                  Yao Lu and
                  Song Han},
  title        = {Long{VILA}: Scaling Long-Context Visual Language Models for Long Videos},
  journal      = {CoRR},
  volume       = {abs/2408.10188},
  year         = {2024},
  url          = {https://doi.org/10.48550/arXiv.2408.10188},
  doi          = {10.48550/ARXIV.2408.10188},
  eprinttype    = {arXiv},
  eprint       = {2408.10188},
  bibsource    = {dblp computer science bibliography, https://dblp.org}
}

@article{DBLP:journals/corr/abs-2502-05173,
  author       = {Xilin Wei and
                  Xiaoran Liu and
                  Yuhang Zang and
                  Xiaoyi Dong and
                  Pan Zhang and
                  Yuhang Cao and
                  Jian Tong and
                  Haodong Duan and
                  Qipeng Guo and
                  Jiaqi Wang and
                  Xipeng Qiu and
                  Dahua Lin},
  title        = {Video{R}o{PE}: What Makes for Good Video Rotary Position Embedding?},
  journal      = {CoRR},
  volume       = {abs/2502.05173},
  year         = {2025},
  url          = {https://doi.org/10.48550/arXiv.2502.05173},
  doi          = {10.48550/ARXIV.2502.05173},
  eprinttype    = {arXiv},
  eprint       = {2502.05173},
  bibsource    = {dblp computer science bibliography, https://dblp.org}
}

@inproceedings{DBLP:conf/cvpr/0004LJJCSSL24,
  author       = {Bo He and
                  Hengduo Li and
                  Young Kyun Jang and
                  Menglin Jia and
                  Xuefei Cao and
                  Ashish Shah and
                  Abhinav Shrivastava and
                  Ser{-}Nam Lim},
  title        = {{MA-LMM:} Memory-Augmented Large Multimodal Model for Long-Term Video
                  Understanding},
  booktitle    = {{IEEE/CVF} Conference on Computer Vision and Pattern Recognition,
                  {CVPR} 2024, Seattle, WA, USA, June 16-22, 2024},
  pages        = {13504--13514},
  publisher    = {{IEEE}},
  year         = {2024},
  url          = {https://doi.org/10.1109/CVPR52733.2024.01282},
  doi          = {10.1109/CVPR52733.2024.01282},
  bibsource    = {dblp computer science bibliography, https://dblp.org}
}

@inproceedings{DBLP:conf/icml/BalazevicSP0KH24,
  author       = {Ivana Balazevic and
                  Yuge Shi and
                  Pinelopi Papalampidi and
                  Rahma Chaabouni and
                  Skanda Koppula and
                  Olivier J. H{\'{e}}naff},
  title        = {Memory Consolidation Enables Long-Context Video Understanding},
  booktitle    = {Forty-first International Conference on Machine Learning, {ICML} 2024,
                  Vienna, Austria, July 21-27, 2024},
  publisher    = {OpenReview.net},
  year         = {2024},
  url          = {https://openreview.net/forum?id=qeFgvVVAJ2},
  bibsource    = {dblp computer science bibliography, https://dblp.org}
}

@article{DBLP:journals/corr/abs-2403-14622,
  author       = {Kumara Kahatapitiya and
                  Kanchana Ranasinghe and
                  Jongwoo Park and
                  Michael S. Ryoo},
  title        = {Language Repository for Long Video Understanding},
  journal      = {CoRR},
  volume       = {abs/2403.14622},
  year         = {2024},
  url          = {https://doi.org/10.48550/arXiv.2403.14622},
  doi          = {10.48550/ARXIV.2403.14622},
  eprinttype    = {arXiv},
  eprint       = {2403.14622},
  bibsource    = {dblp computer science bibliography, https://dblp.org}
}

@inproceedings{DBLP:conf/iclr/YuJWCJZXSZWZS25,
  author       = {Sicheng Yu and
                  Chengkai Jin and
                  Huanyu Wang and
                  Zhenghao Chen and
                  Sheng Jin and
                  Zhongrong Zuo and
                  Xiaolei Xu and
                  Zhenbang Sun and
                  Bingni Zhang and
                  Jiawei Wu and
                  Hao Zhang and
                  Qianru Sun},
  title        = {Frame-{V}oyager: Learning to Query Frames for Video Large Language Models},
  booktitle    = {The Thirteenth International Conference on Learning Representations,
                  {ICLR} 2025, Singapore, April 24-28, 2025},
  publisher    = {OpenReview.net},
  year         = {2025},
  url          = {https://openreview.net/forum?id=LNL7zKvm7e},
  bibsource    = {dblp computer science bibliography, https://dblp.org}
}

@inproceedings{DBLP:conf/eccv/KorbarXTZT24,
  author       = {Bruno Korbar and
                  Yongqin Xian and
                  Alessio Tonioni and
                  Andrew Zisserman and
                  Federico Tombari},
  editor       = {Ales Leonardis and
                  Elisa Ricci and
                  Stefan Roth and
                  Olga Russakovsky and
                  Torsten Sattler and
                  G{\"{u}}l Varol},
  title        = {Text-Conditioned Resampler For Long Form Video Understanding},
  booktitle    = {Computer Vision - {ECCV} 2024 - 18th European Conference, Milan, Italy,
                  September 29-October 4, 2024, Proceedings, Part {LXXXVI}},
  series       = {Lecture Notes in Computer Science},
  volume       = {15144},
  pages        = {271--288},
  publisher    = {Springer},
  year         = {2024},
  url          = {https://doi.org/10.1007/978-3-031-73016-0\_16},
  doi          = {10.1007/978-3-031-73016-0\_16},
  bibsource    = {dblp computer science bibliography, https://dblp.org}
}

@inproceedings{DBLP:conf/iccv/MiechZATLS19,
  author       = {Antoine Miech and
                  Dimitri Zhukov and
                  Jean{-}Baptiste Alayrac and
                  Makarand Tapaswi and
                  Ivan Laptev and
                  Josef Sivic},
  title        = {How{T}o100{M}: Learning a Text-Video Embedding by Watching Hundred Million
                  Narrated Video Clips},
  booktitle    = {2019 {IEEE/CVF} International Conference on Computer Vision, {ICCV}
                  2019, Seoul, Korea (South), October 27 - November 2, 2019},
  pages        = {2630--2640},
  publisher    = {{IEEE}},
  year         = {2019},
  url          = {https://doi.org/10.1109/ICCV.2019.00272},
  doi          = {10.1109/ICCV.2019.00272},
  bibsource    = {dblp computer science bibliography, https://dblp.org}
}

@inproceedings{DBLP:conf/iccv/KrishnaHRFN17,
  author       = {Ranjay Krishna and
                  Kenji Hata and
                  Frederic Ren and
                  Li Fei{-}Fei and
                  Juan Carlos Niebles},
  title        = {Dense-Captioning Events in Videos},
  booktitle    = {{IEEE} International Conference on Computer Vision, {ICCV} 2017, Venice,
                  Italy, October 22-29, 2017},
  pages        = {706--715},
  publisher    = {{IEEE} Computer Society},
  year         = {2017},
  url          = {https://doi.org/10.1109/ICCV.2017.83},
  doi          = {10.1109/ICCV.2017.83},
  bibsource    = {dblp computer science bibliography, https://dblp.org}
}

@inproceedings{DBLP:conf/nips/NarasimhanRD21,
  author       = {Medhini Narasimhan and
                  Anna Rohrbach and
                  Trevor Darrell},
  editor       = {Marc'Aurelio Ranzato and
                  Alina Beygelzimer and
                  Yann N. Dauphin and
                  Percy Liang and
                  Jennifer Wortman Vaughan},
  title        = {{CLIP-I}t! Language-Guided Video Summarization},
  booktitle    = {Advances in Neural Information Processing Systems 34: Annual Conference
                  on Neural Information Processing Systems 2021, NeurIPS 2021, December
                  6-14, 2021, virtual},
  pages        = {13988--14000},
  year         = {2021},
  url          = {https://proceedings.neurips.cc/paper/2021/hash/7503cfacd12053d309b6bed5c89de212-Abstract.html},
  bibsource    = {dblp computer science bibliography, https://dblp.org}
}

@article{DBLP:journals/corr/abs-2503-01201,
  author       = {Louis Mahon and
                  Mirella Lapata},
  title        = {Parameter-free Video Segmentation for Vision and Language Understanding},
  journal      = {CoRR},
  volume       = {abs/2503.01201},
  year         = {2025},
  url          = {https://doi.org/10.48550/arXiv.2503.01201},
  doi          = {10.48550/ARXIV.2503.01201},
  eprinttype    = {arXiv},
  eprint       = {2503.01201},
  bibsource    = {dblp computer science bibliography, https://dblp.org}
}

@article{DBLP:journals/corr/abs-2507-06261,
  author       = {Gheorghe Comanici and
                  Eric Bieber and
                  Mike Schaekermann and
                  Ice Pasupat and
                  Noveen Sachdeva and
                  Inderjit S. Dhillon and
                  Marcel Blistein and
                  Ori Ram and
                  Dan Zhang and
                  Evan Rosen and
                  Luke Marris and
                  Sam Petulla and
                  Colin Gaffney and
                  Asaf Aharoni and
                  Nathan Lintz and
                  Tiago Cardal Pais and
                  Henrik Jacobsson and
                  Idan Szpektor and
                  Nan{-}Jiang Jiang and
                  Krishna Haridasan and
                  Ahmed Omran and
                  Nikunj Saunshi and
                  Dara Bahri and
                  Gaurav Mishra and
                  Eric Chu and
                  Toby Boyd and
                  Brad Hekman and
                  Aaron Parisi and
                  Chaoyi Zhang and
                  Kornraphop Kawintiranon and
                  Tania Bedrax{-}Weiss and
                  Oliver Wang and
                  Ya Xu and
                  Ollie Purkiss and
                  Uri Mendlovic and
                  Ila{\"{\i}} Deutel and
                  Nam Nguyen and
                  Adam Langley and
                  Flip Korn and
                  Lucia Rossazza and
                  Alexandre Ram{\'{e}} and
                  Sagar Waghmare and
                  Helen Miller and
                  Nathan Byrd and
                  Ashrith Sheshan and
                  Raia Hadsell Sangnie Bhardwaj and
                  Pawel Janus and
                  Tero Rissa and
                  Dan Horgan and
                  Sharon Silver and
                  Ayzaan Wahid and
                  Sergey Brin and
                  Yves Raimond and
                  Klemen Kloboves and
                  Cindy Wang and
                  Nitesh Bharadwaj Gundavarapu and
                  Ilia Shumailov and
                  Bo Wang and
                  Mantas Pajarskas and
                  Joe Heyward and
                  Martin Nikoltchev and
                  Maciej Kula and
                  Hao Zhou and
                  Zachary Garrett and
                  Sushant Kafle and
                  Sercan Arik and
                  Ankita Goel and
                  Mingyao Yang and
                  Jiho Park and
                  Koji Kojima and
                  Parsa Mahmoudieh and
                  Koray Kavukcuoglu and
                  Grace Chen and
                  Doug Fritz and
                  Anton Bulyenov and
                  Sudeshna Roy and
                  Dimitris Paparas and
                  Hadar Shemtov and
                  Bo{-}Juen Chen and
                  Robin Strudel and
                  David Reitter and
                  Aurko Roy and
                  Andrey Vlasov and
                  Changwan Ryu and
                  Chas Leichner and
                  Haichuan Yang and
                  Zelda Mariet and
                  Denis Vnukov and
                  Tim Sohn and
                  Amy Stuart and
                  Wei Liang and
                  Minmin Chen and
                  Praynaa Rawlani and
                  Christy Koh and
                  JD Co{-}Reyes and
                  Guangda Lai and
                  Praseem Banzal and
                  Dimitrios Vytiniotis and
                  Jieru Mei and
                  Mu Cai},
  title        = {Gemini 2.5: Pushing the Frontier with Advanced Reasoning, Multimodality,
                  Long Context, and Next Generation Agentic Capabilities},
  journal      = {CoRR},
  volume       = {abs/2507.06261},
  year         = {2025},
  url          = {https://doi.org/10.48550/arXiv.2507.06261},
  doi          = {10.48550/ARXIV.2507.06261},
  eprinttype    = {arXiv},
  eprint       = {2507.06261},
  bibsource    = {dblp computer science bibliography, https://dblp.org}
}

@article{DBLP:journals/corr/abs-2503-20215,
  author       = {Jin Xu and
                  Zhifang Guo and
                  Jinzheng He and
                  Hangrui Hu and
                  Ting He and
                  Shuai Bai and
                  Keqin Chen and
                  Jialin Wang and
                  Yang Fan and
                  Kai Dang and
                  Bin Zhang and
                  Xiong Wang and
                  Yunfei Chu and
                  Junyang Lin},
  title        = {Qwen2.5-Omni Technical Report},
  journal      = {CoRR},
  volume       = {abs/2503.20215},
  year         = {2025},
  url          = {https://doi.org/10.48550/arXiv.2503.20215},
  doi          = {10.48550/ARXIV.2503.20215},
  eprinttype    = {arXiv},
  eprint       = {2503.20215},
  bibsource    = {dblp computer science bibliography, https://dblp.org}
}

@article{DBLP:journals/corr/abs-2412-15115,
  author       = {An Yang and
                  Baosong Yang and
                  Beichen Zhang and
                  Binyuan Hui and
                  Bo Zheng and
                  Bowen Yu and
                  Chengyuan Li and
                  Dayiheng Liu and
                  Fei Huang and
                  Haoran Wei and
                  Huan Lin and
                  Jian Yang and
                  Jianhong Tu and
                  Jianwei Zhang and
                  Jianxin Yang and
                  Jiaxi Yang and
                  Jingren Zhou and
                  Junyang Lin and
                  Kai Dang and
                  Keming Lu and
                  Keqin Bao and
                  Kexin Yang and
                  Le Yu and
                  Mei Li and
                  Mingfeng Xue and
                  Pei Zhang and
                  Qin Zhu and
                  Rui Men and
                  Runji Lin and
                  Tianhao Li and
                  Tingyu Xia and
                  Xingzhang Ren and
                  Xuancheng Ren and
                  Yang Fan and
                  Yang Su and
                  Yichang Zhang and
                  Yu Wan and
                  Yuqiong Liu and
                  Zeyu Cui and
                  Zhenru Zhang and
                  Zihan Qiu},
  title        = {Qwen2.5 Technical Report},
  journal      = {CoRR},
  volume       = {abs/2412.15115},
  year         = {2024},
  url          = {https://doi.org/10.48550/arXiv.2412.15115},
  doi          = {10.48550/ARXIV.2412.15115},
  eprinttype    = {arXiv},
  eprint       = {2412.15115},
  bibsource    = {dblp computer science bibliography, https://dblp.org}
}
\bibliographystyle{acl_natbib}

\appendix

\section{Prompts}

\subsection{Clip Selection}
\label{app:clip_selection}

\subsubsection{Clip Selection Prompt}

We provide below the prompt for the clip selection with an LLM.

\begin{itemize}
\item {\color{blue} MOVIE\_NAME} is the movie title.
\item {\color{red} <CAPTIONS>} refers to all the captions generated for the 20-second video clips using the lightweight captioning model (Qwen2.5-Omni).
\item {\color{blue} NB\_CAPTIONS} is the number of selected clips~(same as~$K$).
\end{itemize}

\begin{tcolorbox}[breakable, colback=white, colframe=green!20, left=2pt,  coltitle=black,  title=\textbf{}]

Here are captions from the movie {\color{blue} MOVIE\_NAME}:\\

{\color{red} <CAPTIONS>}\\

What are the {\color{blue} NB\_CAPTIONS} most important Captions that describe important action or visual event you would include in the existing Summary of the movie {\color{blue} MOVIE\_NAME}?\\
Provide your answer in the following way:\\
1. Caption caption\_number: Justification why the Caption describes crucial action for the summary\\
2. Caption caption\_number: Justification why the Caption describes crucial action for the summary\\

...\\

{\color{blue} NB\_CAPTIONS}. Caption caption\_number: Justification why the Caption describes crucial action for the summary\\

Answer:

\end{tcolorbox}

\subsubsection{Two-shot Clip Selection Examples}

We annotate and use the following few-shot examples for the clip selection task. Those examples are derived from the movies \textit{Forrest Gump (1994)} and \textit{Wonder Woman (2017)}.

\begin{tcolorbox}[breakable, colback=white, colframe=green!20, left=2pt,  coltitle=black,  title=\textbf{}]

Here are captions from the movie Forrest Gump:\\

Caption 1110000: In the video, a man and woman sit on a bench in a park. The man is wearing a suit and tie while the woman wears casual clothes. They appear to be reading books together as they sit side by side. The man then turns his attention towards the woman and starts talking about something. He mentions that life is like a box of chocolates and you never know what you're going to get. He also comments on how comfortable her shoes must be and suggests she could walk all day in them.

Caption 1130000: Forrest is sitting on a bench outside. He then sits inside a doctor's office with his legs up on the table. The doctor removes Forrest's leg braces and asks him to stand up. Forrest stands up and walks around the room.

Caption 1150000: The dialogue reveals that the woman is explaining the origin of the character\'s name "Forrest Gump." She mentions that the "Forrest" part of the name comes from an incident where they were related to someone who started a club called the Ku Klux Klan. The woman explains that the "Gump" part of the name was given because sometimes people do things that don\'t make sense.

Caption 1170000: The video shows a group of boys chasing Forrest Gump as he runs down a dirt road. The boys are shouting at him to run faster, while Forrest continues to run without looking back. One of the boys falls over, but gets up quickly and continues chasing Forrest. The other boys also catch up with Forrest and start to chase him more aggressively. As they get closer, one of the boys throws a rock at Forrest, who ducks to avoid it. Another boy tries to kick him, but misses. The boys continue to chase Forrest until he reaches his home, where his mother is waiting for him. She tells him that miracles happen every day, and that some people may not believe them, but they still exist.

Caption 1190000: The man is running on the field, and he jumps over the fence. He runs to the football field and throws the ball. The coaches are watching him.

Caption 1210000: The video shows a scene where a woman holding a baby sits on a bench next to another woman who is reading a book. A man in a suit is sitting on the other side of the bench with his suitcase beside him. The woman with the baby stands up and walks away from the bench while talking to the man. She then sits back down on the bench and continues talking to him. In the background, there is a bus passing by. The dialogue includes the woman asking if the bus is the number nine, but the man corrects her and says it's the number four. They also have a conversation about someone named Wallace getting shot while they were in college.

Caption 1230000: The video shows a woman reading a book to her son on their bed. The boy asks his mother about vacation, and she explains that it is when someone goes somewhere and never comes back. \\

What are the 3 most important Captions that describe important action or visual event you would include in a Summary of the movie Forrest Gump? \\
Provide your answer in the following way:\\
1. Caption caption\_number: Justification why the Caption describes crucial action for the summary\\
2. Caption caption\_number: Justification why the Caption describes crucial action for the summary\\
3. Caption caption\_number: Justification why the Caption describes crucial action for the summary\\

Answer:\\
Caption 1130000: Justification: This caption depicts the removal of Forrest's leg braces, a pivotal moment signifying his physical transformation and newfound freedom. \\
Caption 1170000: Justification: This caption illustrates the bullying Forrest faces and his eventual discovery of his running ability, a recurring motif in the film.\\
Caption 1190000: Justification: This caption depicts Forrest's accidental entry into the world of football, showcasing his unexpected athletic talent.\\

\end{tcolorbox}

\begin{tcolorbox}[breakable, colback=white, colframe=green!20, left=2pt,  coltitle=black,  title=\textbf{}]

Here are captions from the movie Wonder Woman:\\

Caption 4210000: The scene opens with a man sitting at his desk, looking at his watch. He then turns to face another man standing before him. The man in uniform speaks to the other man, telling him that he will do nothing. The man in uniform then walks away as the other man looks on. The scene ends with the man in uniform walking out of the room.\\
Caption 4230000: Diana and Steve are walking down the stairs. Steve is talking to Diana. Steve is angry at Diana for not fighting back against Ares. He tells her that she didn't stand her ground because there was no chance of changing Ares' mind. He also tells her that millions of people will die if they don't fight back. He tells her that his people are next. Summary: Steve is angry at Diana for not fighting back against Ares. He tells her that she didn't stand her ground because there was no chance of changing Ares' mind. He also tells her that millions of people will die if they don't fight back. He tells her that his people are next.\\
Caption 4250000: The video shows a man sitting on a chair in a room. A bomb is thrown into the room and explodes. The man gets up and runs out of the door. He then talks to another man who is standing outside the door. The man inside the room is coughing and choking on smoke.\\

What are the 1 most important Captions that describe important action or visual event you would include in a Summary of the movie Wonder Woman?\\
Provide your answer in the following way:\\
1. Caption caption\_number: Justification why the Caption describes crucial action for the summary\\

Answer:\\
Caption 4250000: Justification: This caption depicts a sudden and violent attack, showcasing the dangers faced by the characters and the chaos of the war. It emphasizes the element of surprise and the characters' ability to react quickly to threats. Therefore the Caption depicts important visual action of event.\\

\end{tcolorbox}

\subsection{Clip captioning Prompts} \label{subsec:clip_captioning_prompt}

Below are the prompt templates used for the lightweight captioning with Qwen2.5-Omni and the recaptioning with Gemini 2.5 Flash-Lite. The video clips are processed by both VLMs at one frame per second (1 fps) and including the audio.

\subsubsection{Lightweight Captioning with Qwen2.5-Omni}

\begin{tcolorbox}[colback=white, colframe=green!20, left=2pt,  coltitle=black,  title=\textbf{}]

{\color{red} <VIDEO CLIP (1 fps)+ AUDIO>}\\

Describe both the action and Summarize the corresponding dialogue.

\end{tcolorbox}

\subsubsection{Recaptioning with Gemini 2.5 Flash-Lite} \label{app:clip_caption_no_char}

\begin{tcolorbox}[colback=white, colframe=green!20, left=2pt,  coltitle=black,  title=\textbf{}]

{\color{red} <VIDEO CLIP (1 fps)+ AUDIO>}\\

Describe both the video, action and dialogue in one paragraph

\end{tcolorbox}

\subsection{Summarization Prompt} \label{subsec:summarization_prompt}

We provide here the prompt we used for generating multimodal summaries in all our experiments.

We explicitly state in our prompt that the produced summary has to be multimodal by including both relevant visual and textual elements from either the transcript lines and the video captions.

We fix the generated summary length to~1000 words in the prompt and truncate the output beyond that limit. Note that the average summary length of the groundtruth summaries in the whole MovieSum dataset (train and test sets) is 635 words.

\begin{tcolorbox}[colback=white, colframe=green!20, left=2pt,  coltitle=black,  title=\textbf{}]

{\color{red} <TRANSCRIPT> or <SCREENPLAY>}\\

Generate a comprehensive multimodal summary of exactly 1000 words of the movie based on the provided dialogue and the most important visual elements.

Your summary should:

Synthesize information from both the dialogue (transcript) and the important visual events (visual analysis).

Your overall summary should contain exactly 1000 words. Do not refer to external websites, movie databases or plot summaries.

\end{tcolorbox}

\subsection{Clip Selection Reference}
\label{subsec:reference_clips}

\subsubsection{Fact Identification}

We provide below the prompt for extracting all the facts from the groundtruth summary by first splitting the summary into sentences and then each sentence into facts.

\begin{tcolorbox}[breakable, colback=white, colframe=green!20, left=2pt,  coltitle=black,  title=\textbf{}]
Summary:\\
{\color{red} <SUMMARY>}\\

For every sentence from the Summary, decompose the sentence in a list of facts (at least 1). Each fact can be only part of a sentence and should convey a single piece of information about the story.\\

Example:

Sentence 1:\\
*\\

Sentence 2:\\
*\\
...\\

Sentence N:\\
*
\end{tcolorbox}

\subsubsection{Visual Fact Classification} \label{subsubsec:visual_fact_prompt}

Given the gold screenplay of a movie, we are able to infer which groundtruth summary fact is~\texttt{Visual} or~\texttt{Textual}.

We prompt an LLM in zero-shot to quote the line from the screenplay that supports a given groundtruth summary fact. If the quoted line belongs to the dialogue, then the fact is classified as~\texttt{Textual}. Otherwise, if it corresponds to a clip caption, then the fact is classified as~\texttt{Visual}. We provide below the prompt being used for the task of visual fact classification.

\begin{tcolorbox}[colback=white, colframe=green!20, left=2pt,  coltitle=black,  title=\textbf{Fact Evaluation against Transcripts}]
Screenplay:\\
{\color{red} <SCREENPLAY>}\\

For every fact below:\\
{\color{red}<ALL FACTS>}

-> Find the information in the above Screenplay. Quote a line from the Screenplay.\\

Example:

Fact 1: Recopy the Fact\\
-> Quoted line from Screenplay\\

Fact 2: Recopy the Fact\\
-> Quoted line from Screenplay\\
...\\

Fact N: Recopy the Fact\\
-> Quoted line from Screenplay
\end{tcolorbox}

\subsection{MFactSum evaluation} \label{subsec:mfactsum_prompt}

We present below the prompt used to evaluate the visual or textual recall of groundtruth summary facts. Specifically, this prompt tests whether each groundtruth fact is supported by the predicted summary.

\begin{tcolorbox}[colback=white, colframe=green!20, left=2pt,  coltitle=black,  title=\textbf{}]
Summary:\\
{\color{red} <SUMMARY>}\\

Task:\\
For each fact listed below, determine whether the exact meaning of the fact is explicitly present in the summary above.\\

Instructions:\\
You must justify your answer by quoting or paraphrasing the relevant part of the summary.
If the fact is not explicitly present, even if it seems implied or suggested, you must answer No.\\
Do not accept facts just because they are likely, inferable, or assumed from context.
However, do allow for reasonable paraphrasing or rewording. If the summary conveys the same meaning as the fact using different but equivalent words, answer Yes.\\

Format:\\

Fact 1: [Recopy the Fact]\\
1. Justification (quote or paraphrase from the summary, and explain how it matches the fact)\\
2. Yes\\

Fact 2: [Recopy the Fact]\\
1. Justification\\
2. No\\
...\\

Fact N: [Recopy the Fact]\\
1. Justification\\
2. Yes\\

List of all Facts:\\
{\color{red} <ALL FACTS>}\\

\end{tcolorbox}

\onecolumn

\section{Additional Experiments} \label{app:experiments}

In Tables~\ref{tab:main-results-MFactSum_15} and~\ref{tab:main-results-MFactSum_qwen}, we report the results using respectively Gemini 1.5 Flash and Qwen2.5-72B-Instruct in place of Gemini 2.5 Flash as the summarization model in our pipeline~(Figure~\ref{fig:pipeline}).

\begin{table*}[htbp]
\center
\setlength{\tabcolsep}{2pt}
\resizebox{0.75\textwidth}{!}{
 \begin{tabular}{@{}cccc|ccc|c@{}}
\toprule & vis-rec & text-rec & \textsc{MFS} & r1 & r2 & rlsum & METEOR \\
\midrule
\multicolumn{1}{l}{\textbf{Transcripts (no video)}} & 13.17 & 18.41 & 15.79 & 34.19 & 7.10 & 32.64 & 26.52 \\
\noalign{\vskip 0.5ex}
\cdashline{1-8}[2pt/2pt]
\noalign{\vskip 0.5ex}
\multicolumn{1}{c}{\textbf{Built Screenplay (50 clips)}} & & & & & & & \\
\multicolumn{1}{l}{random clips} & 14.20 & 18.68 & 16.44 & 33.79 & 7.12 & 32.13 & 26.78 \\
\multicolumn{1}{l}{silent clips} & 14.11 & 19.54 & 16.83 & \textbf{34.80} & \textbf{7.41} & \textbf{33.15} & 27.16 \\
\multicolumn{1}{l}{\textit{our clips zero-shot (Qwen2.5-Omni-7B)}} & 14.88 & 19.72 & 17.30 & 33.82 & 7.14 & 32.17 & 27.15 \\
\multicolumn{1}{l}{\textit{our clips two-shot (Qwen2.5-Omni-7B)}} & \textbf{16.88} & \textbf{20.00} & \textbf{18.44} & 34.25 & 7.40 & 32.57 & \textbf{27.45} \\
\noalign{\vskip 0.5ex}
\cdashline{1-8}[2pt/2pt]
\noalign{\vskip 0.5ex}
\multicolumn{1}{l}{\textbf{Built Screenplay (reference clips)}} & 16.45 & 19.04 & 17.75 & 34.86 & 7.37 & 33.14 & 27.21 \\
\midrule
\multicolumn{1}{l}{\textbf{Gold Screenplay}} & 22.78 & 23.07 & 22.92 & 34.87 & 7.80 & 33.03 & 28.41 \\

\bottomrule
\end{tabular}
}
\caption{\textbf{Evaluation results using Gemini 1.5 Flash for summarization.} Evaluations are made on the MovieSum test set.
Column descriptions are the same as in Table~\ref{tab:motivation}. Best results are in \textbf{bold}.} \label{tab:main-results-MFactSum_15}
\end{table*}

\begin{table*}[htbp]
\center
\setlength{\tabcolsep}{2pt}
\resizebox{0.75\textwidth}{!}{
 \begin{tabular}{@{}cccc|ccc|c@{}}
\toprule & vis-rec & text-rec & \textsc{MFS} & r1 & r2 & rlsum & METEOR \\
\midrule
\multicolumn{1}{l}{\textbf{Transcripts (no video)}} & 17.27 & 23.92 & 20.59 & 41.88 & 10.41 & 40.08 & 29.88 \\
\noalign{\vskip 0.5ex}
\cdashline{1-8}[2pt/2pt]
\noalign{\vskip 0.5ex}
\multicolumn{1}{c}{\textbf{Built Screenplay (50 clips)}} & & & & & & & \\
\multicolumn{1}{l}{random clips} & 17.69 & 24.04 & 20.86 & 41.80 & 10.45 & 39.88 & 29.80 \\
\multicolumn{1}{l}{silent clips} & 18.56 & 24.28 & 21.42 & \textbf{42.20} & \textbf{10.66} & \textbf{40.10} & \textbf{30.16} \\
\multicolumn{1}{l}{\textit{our clips zero-shot (Qwen2.5-Omni-7B)}} & \textbf{19.25} & \textbf{24.32} & \textbf{21.79} & 41.79 & 10.58 & 39.72 & 29.77 \\
\multicolumn{1}{l}{\textit{our clips two-shot (Qwen2.5-Omni-7B)}} & 18.71 & 24.08 & 21.39 & 41.66 & 10.48 & 39.81 & 29.56 \\
\noalign{\vskip 0.5ex}
\cdashline{1-8}[2pt/2pt]
\noalign{\vskip 0.5ex}
\multicolumn{1}{l}{\textbf{Built Screenplay (reference clips)}} & 19.44 & 23.62 & 21.53 & 42.15 & 10.70 & 40.06 & 30.01 \\
\midrule
\multicolumn{1}{l}{\textbf{Gold Screenplay}} & 28.77 & 27.81 & 28.29 & 43.55 & 11.32 & 41.43 & 31.47 \\

\bottomrule
\end{tabular}
}
\caption{\textbf{Evaluation results using Qwen2.5-72B-Instruct for summarization.} Evaluations are made on the MovieSum test set.
Column descriptions are the same as in Table~\ref{tab:motivation}. Best results are in \textbf{bold}.} \label{tab:main-results-MFactSum_qwen}
\end{table*}

\section{Human Evaluation of the Clip Selection Reference} \label{app:human_eval_clip}

We report the results of our human evaluation against the first annotator in Table~\ref{tab:human_clip_selection}.

\begin{table*}[htbp]
\small
\center
\setlength{\tabcolsep}{2pt}
\resizebox{0.9\textwidth}{!}{
 \begin{tabular}{@{}lcccc|c@{}}
\toprule & Shining (1980) & Dark Knight (2008) & Imitation Game (2014) & Black Panther (2018) & Average/Total \\
\midrule
Precision & 80.6 & 72.2 & 81.8 & 100.0 & 83.65 \\
Recall & 89.5 & 90.0 & 91.7 & 90.0 & 90.3 \\
F1 Score & 84.8 & 80.1 & 86.5 & 94.7 & 86.5 \\
\midrule
Nb reference clips & 31 & 54 & 11 & 12 & 108 \\
\bottomrule
\end{tabular}
}
\caption{\textbf{Human evaluation of the clip selection reference by the first annotator.} We report the Precision, Recall and F1 scores between the clip selection reference~(see Section~\ref{subsec:pseudo_clips}) and the human reference on all 4 movies. We also report the number of clips in the clip selection reference for each movie.} \label{tab:human_clip_selection}
\end{table*}

\end{document}